\documentclass[letterpaper]{article} %
\usepackage[preprint]{aaai2027}
\usepackage[hyphens]{url}  %
\usepackage{graphicx} %
\usepackage{natbib}  %
\usepackage{caption} %
\usepackage{amsmath}
\usepackage{amssymb}
\usepackage{algorithm}
\usepackage{algpseudocode}
\usepackage{booktabs}
\usepackage{colortbl}
\usepackage{pifont}
\usepackage{subcaption}
\usepackage{tikz}
\usepackage{xspace}
\usepackage{cleveref}
\usetikzlibrary{positioning,arrows.meta}

\crefname{figure}{Figure}{Figures}
\crefname{table}{Table}{Tables}
\crefname{section}{Section}{Sections}
\crefname{subsection}{Section}{Sections}
\crefname{subsubsection}{Section}{Sections}
\crefname{equation}{Equation}{Equations}
\crefname{algorithm}{Algorithm}{Algorithms}

\newcommand{\xmark}{\ding{55}}

\colorlet{bestclr}{teal!60}
\colorlet{secondclr}{teal!40}
\colorlet{thirdclr}{teal!20}
\definecolor{VioletRed}{RGB}{199,21,133}
\newcommand{\firstplace}[1]{\textbf{#1}}
\newcommand{\secondplace}[1]{\underline{#1}}

\def\tabclrscheme{The best and second-best result in each column are shown in bold and underlined, respectively.}
\def\tabclrschemefollow{Ranking formatting follows~\cref{tab:exp_non_isometric}.}

\DeclareRobustCommand{\eg}{\emph{e.g.}\@\xspace}

\newcommand{\ourmethod}{GAMMA\xspace}

\newcommand{\affil}[1]{\textsuperscript{\rm #1}}

\title{Geometry-Grounded Articulation and Test-Time Adaptation for Zero-Shot, Scalable Shape Matching in the Wild}
\author{
    Qilong Liu\equalcontrib\affil{1},
    Qinfeng Xiao\equalcontrib\affil{1},
    Chenyuan Yi\affil{1},
    Yongsheng Lin\affil{1},
    Liying Zhang\affil{1},
    Kit-lun Yick\corresponding\affil{1,2}
}
\affiliations{
    \affil{1}Hong Kong Polytechnic University, HK SAR\\
    \affil{2}Artificial Intelligence in Design, HK SAR\\
}

\begin{document}

\pagestyle{plain}
\maketitle
\thispagestyle{plain}

\begin{abstract}

Dense correspondence on in-the-wild 3D scans must handle severe non-isometric deformation, partial observations, topology artifacts, irregular discretization, and extreme resolution. Together, these conditions make shape matching both a robustness and scalability problem. Existing methods typically handle only part of this regime and often require expensive spectral or dense feature processing, repeated pairwise optimization, or dataset-specific training. To address the above issues, we present \textbf{\ourmethod}, a zero-shot framework based on \emph{Geometry-Grounded Articulation}. For each scan, \ourmethod renders multiple views and uses a pretrained estimator with a category-specific parametric model to produce one template estimate per view. Because these estimates share a fixed vertex ordering, confidence-weighted consensus fuses them into one canonical template and establishes sparse matches across scans. We refine these matches on the input geometry to recover accurate dense correspondences. All core matching steps run at inference time and require neither target-dataset correspondence training nor costly Laplace--Beltrami operator (LBO) computation on the input scans. Each fused template is cached, so additional pair queries reuse the canonicalization. \ourmethod obtains average geodesic errors of \textbf{2.9} on TOPKIDS and \textbf{2.4} on remeshed SMAL, reducing error relative to URSSM by 67\% and 60\%, respectively. Complementary evaluations report lower aggregate errors than the evaluated baselines on BeCoS-processed partial subsets and demonstrate stable matching accuracy from decimated FAUST meshes to raw scans with 160k--200k vertices. These results demonstrate accurate correspondence across complementary settings and collection-level scalability after per-shape caching.

\end{abstract}

\begin{links}
  \link{Code}{https://github.com/liu-qilong/GAMMA}
  \link{Project}{https://liu-qilong.github.io/GAMMA}
\end{links}

\begin{figure*}[!t]
    \centering
    \includegraphics[width=\linewidth]{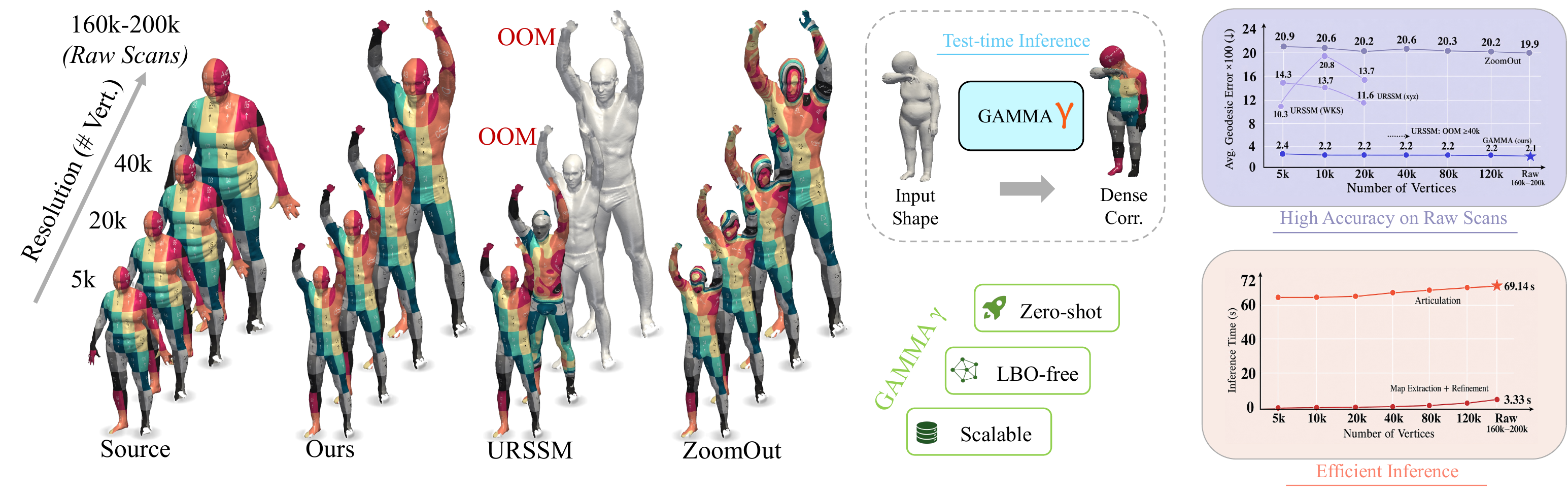}
    \caption{\textbf{Zero-shot dense correspondence on high-resolution raw scans.} \textit{Left:} Selected FAUST transfers remain coherent from 5k vertices to the original 160k--200k scans. In the archived implementation, URSSM encounters memory failure from 40k onward, while chunked ZoomOut produces high-error maps. \textit{Center:} \ourmethod grounds pretrained category priors through per-shape test-time inference, requiring neither target-benchmark correspondence training nor an input LBO eigenbasis. \textit{Right:} The template-domain error remains low across resolutions; articulation is measured per shape, whereas map composition and ZoomOut are measured per pair.}
    \label{fig:teaser}
\end{figure*}

\section{Introduction}

Shape matching aims to establish dense correspondences between 3D shapes and enables a wide range of applications, \eg, texture transfer~\cite{ezuz2017deblurring}, shape interpolation~\cite{eisenberger2021neuromorph,cao2024spectral}, statistical body modeling~\cite{maheshwari2023transfer4d}, and robotic manipulation~\cite{zhu2025densematcher}. Most correspondence benchmarks, however, are built from clean, complete, small-scale meshes. We instead target \emph{in-the-wild shape matching}, where a high-resolution raw scan can combine severe non-isometric deformation with missing regions, fused parts, self-intersections, irregular discretization, and hundreds of thousands of vertices. These factors reinforce one another: articulation weakens local geometric cues, acquisition and topological artifacts corrupt connectivity, and extreme resolution magnifies the cost of dense or spectral computation. Matching in the wild is thus not only a robustness problem, but also a fundamental challenge of scalability and computational efficiency.

Existing methods provide strong solutions to individual parts of this problem, yet their cost and assumption profiles leave the full conjunction largely unresolved. Functional-map methods~\cite{ovsjanikov2012functional,litany2017deep,donati2020deep} offer compact correspondence representations, and recent variants improve their memory scaling~\cite{magnet2024memory}; nevertheless, the standard pipeline constructs descriptors and maps from an input Laplace--Beltrami operator (LBO) eigenbasis, whose cost and stability can become problematic on large, irregular, or topologically corrupted scans. Learned template and correspondence methods~\cite{groueix20183d,zheng2021deep,zhang2023self} exploit category structure effectively, but typically require correspondence supervision or training on shape collections. Semantic cues improve transfer, although their costs are paid elsewhere: ZSC~\cite{abdelreheem2023zero} optimizes each shape pair, Diff3F~\cite{dutt2024diffusion} extracts dense image features, and recent dense semantic matchers~\cite{zhu2025densematcher,xiao2026universal,xiao2026glass} use category-specific training. Methods designed explicitly for high-resolution matching~\cite{amrani2025high} further demonstrate progress in scalability. What remains missing is a way to ground available articulation priors once per scan and reuse the result across pairs, without target-dataset correspondence training or dependence on a spectral basis.

We introduce \emph{\textbf{\ourmethod}}, a zero-shot framework based on \emph{Geometry-Grounded Articulation}. Here, \emph{zero-shot} means that no correspondence model is trained or adapted on the target benchmark. \ourmethod uses a category-specific parametric model and a compatible pretrained estimator, and optimizes only per-shape geometry at inference time. For each scan, \ourmethod renders multiple views, and the estimator predicts one parametric template per view. The scan geometry then aligns and refines these predictions while filtering inconsistent observations. Because all templates share a fixed vertex order, confidence-weighted consensus fuses them into one canonical template and provides sparse correspondences across scans. We refine these correspondences on the original geometry to recover accurate dense point-wise correspondences without computing an LBO on either input (time-consuming for high-resolution scans). Processing each raw scan still depends on its resolution, but the resulting template and sparse matches are cached once per scan, so additional pair queries reuse the inference and optimization.

We evaluate these claims across settings involving strong non-isometry, partial observations, and raw-scan artifacts. On TOPKIDS and remeshed SMAL, \ourmethod obtains mean geodesic errors (\(\times100\)) of 2.9 and 2.4, reducing error relative to URSSM~\cite{cao2023unsupervised} by 67\% and 60\%, respectively. On BeCoS-processed FAUST and KIDS partial subsets, it reports lower aggregate errors than the evaluated baselines, while the raw-FAUST experiment demonstrates operation on 160k--200k-vertex scans. In summary, our main contributions are as follows:

\begin{itemize}
    \item \textbf{Problem formulation.} We formulate in-the-wild shape matching as a coupled robustness-and-scalability problem for high-resolution raw scans. Our prior-supported zero-shot setting uses compatible pretrained category priors without target-benchmark training or adaptation; only per-shape geometric variables are optimized at inference.
    \item \textbf{Per-shape grounding.} We introduce Geometry-Grounded Articulation, which uses each scan's geometry to reconcile ambiguous multi-view template predictions into a reusable 3D proxy with shared vertex indices.
    \item \textbf{Reusable pairwise matching.} We obtain dense point-to-point maps by sparsely mapping each scan to the shared indices and composing the assignments. The primary matcher computes no input LBO eigenbasis; reliability and sampling-density terms, an unmatched state, and reciprocal filtering provide explicit mechanisms for irregular and partial scans, while caching avoids repeated grounding across pairs.
    \item \textbf{Complementary validation.} We demonstrate high accuracy under strong non-isometry, lower reported aggregate error on the evaluated BeCoS-processed partial subsets, and operation on original 160k--200k-vertex FAUST scans with acquisition and topology artifacts.
\end{itemize}

\section{Related Work}
\label{sec:relatedwork}

\subsection{Robust and Scalable Shape Matching}

Functional maps~\cite{ovsjanikov2012functional} represent correspondence in compact LBO eigenbases; supervised~\cite{litany2017deep,donati2020deep} and unsupervised~\cite{halimi2019unsupervised,roufosse2019unsupervised,cao2023unsupervised} variants learn descriptors and map constraints, ZoomOut~\cite{melzi2019zoomout} progressively raises spectral resolution, and recent work reduces training memory~\cite{magnet2024memory}, improves topology-noise robustness~\cite{cao2024synchronous}, learns task-adapted bases~\cite{luo2026advanced}, or combines rotation-invariant features with complex functional maps~\cite{gao2026rino}. These pipelines require an input LBO; Fast MRF~\cite{roetzer2026fastmrf} instead optimizes discrete triangle assignments per pair. EchoMatch estimates overlap~\cite{xie2025echomatch}, while other work transfers pretrained image features~\cite{abdelreheem2023zero,dutt2024diffusion}, learns neural surface representations~\cite{morreale2024neural,du2025hierarchical} or category-level correspondence~\cite{zhu2025densematcher}, and combines semantic-geometric cues~\cite{yoon2026sgsoft}. SuPa-Match~\cite{amrani2025high} obtains globally consistent assignments via pair-specific optimization, demonstrated up to 10k triangles. These partial, semantic, and high-resolution approaches often require dense feature extraction, category-level training, or repeated pairwise computation. To address these axes jointly in raw scans, \ourmethod pays its main inference and fitting cost per shape and builds the primary map without LBO; cached proxies and sparse factors serve later pairs, while per-scan work remains resolution-dependent.

\subsection{Canonical Priors for Correspondence}

Canonicalization maps to shared domains: 3D-CODED~\cite{groueix20183d} deforms a common template, while DIF, DIT, and DeformShape~\cite{deng2021deformed,zheng2021deep,zhang2023self} learn canonical implicit representations, typically from category-level collections. Fixed-topology human models---SMPL, SMPL-X, STAR, and MHR~\cite{loper2023smpl,pavlakos2019expressive,osman2020star,ferguson2025mhr}---and animal model SMAL~\cite{zuffi20173d} share vertex order across instances, providing semantic indices once observations are grounded. Optimization-based methods fit them to image or 3D evidence~\cite{bogo2016keep,pavlakos2022human,zuffi2018lions}; pretrained image estimators~\cite{yang2025sam3dbody,li2024learning,lyu2025animer,yu2026prima} predict instances. \ourmethod uses an existing parametric model and frozen pretrained estimator; multi-view predictions initialize fitting of only per-scan geometric variables to the observed surface; fusion yields a cached canonical-indexed proxy. No correspondence network is trained or adapted on the target benchmark.

\begin{figure*}[t]
    \centering
    \includegraphics[width=1.0\textwidth]{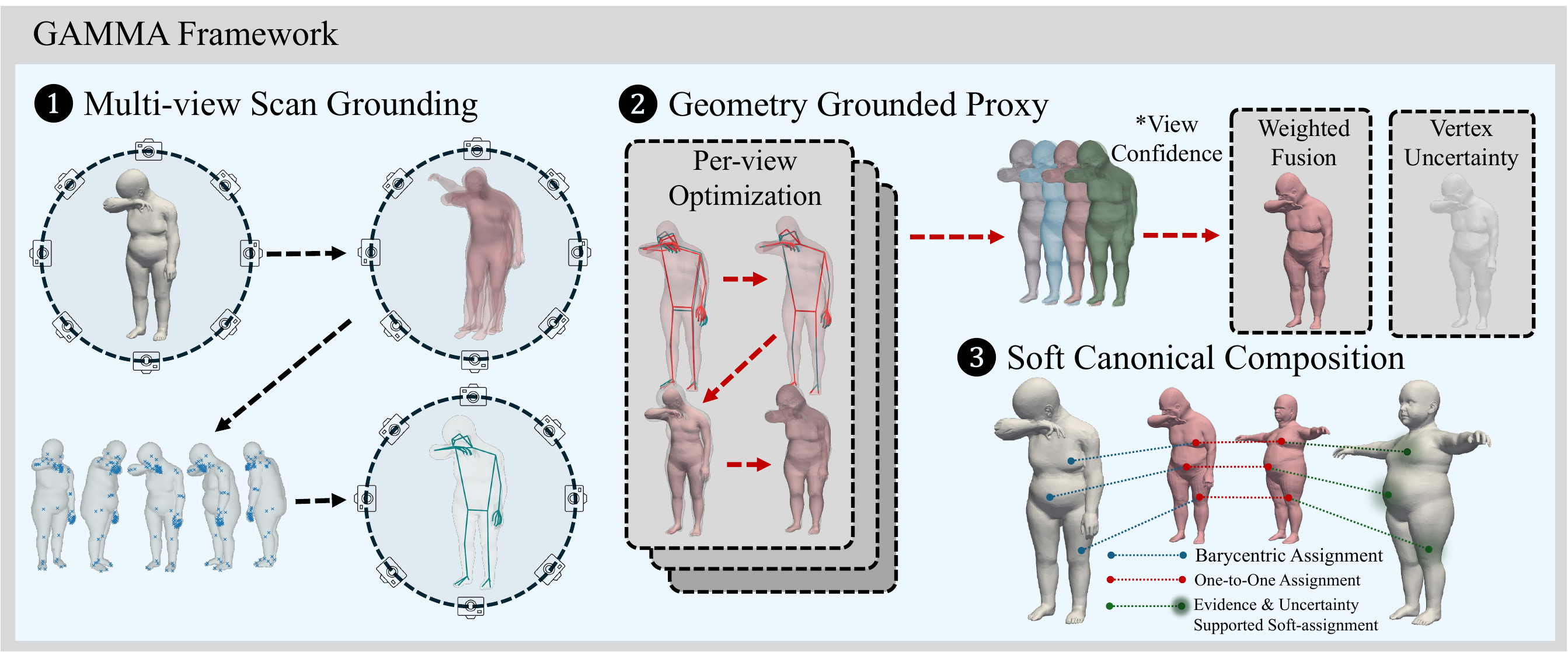}
    \caption{\textbf{Overview of \ourmethod.}
    \textbf{(1) Multi-view scan grounding:} A frozen monocular estimator predicts parametric proposals from rendered scan views; depth-based alignment and cross-view joint triangulation place surface and skeletal evidence in the scan frame.
    \textbf{(2) Geometry-grounded proxy:} Retained proposals are optimized against this evidence, weighted by view confidence, and fused vertex-wise into a fixed-index proxy with empirical per-vertex uncertainty.
    \textbf{(3) Soft canonical composition:} Sparse barycentric scan--proxy assignments are composed through shared canonical indices, while target evidence and uncertainty guide the soft source-to-target correspondence.
    The first two stages are cached per shape; pairwise matching requires only sparse composition, with optional spectral refinement.}
    \label{fig:framework}
\end{figure*}

\section{Method}
\label{sec:methodology}

\Cref{fig:framework} gives an overview of \ourmethod, which separates reusable per-shape processing from per-pair matching. \Cref{sec:prelim} defines the category prior and frozen estimator; \cref{sec:mv-grounding} converts multi-view predictions into scan-frame evidence; \cref{sec:canonization} fits and fuses this evidence into a fixed-index proxy; \cref{sec:dense-corr} composes sparse scan--proxy assignments into the pairwise map; and \cref{sec:scalability} describes caching, query cost, and optional refinement. Monocular inference and optimization therefore run once per shape, while pair queries reuse the cached outputs. Additional notation and derivations appear in~\cref{sec:suppl_method_details,tab:notation}.

\subsection{Preliminaries}
\label{sec:prelim}

Let the source and target shapes be \(\mathcal X=(X,F_X)\) and \(\mathcal Y=(Y,F_Y)\), with raw vertices \(X\in\mathbb R^{N_X\times3}\) and \(Y\in\mathbb R^{N_Y\times3}\), respectively. $F_X, F_Y$ denotes the connectivity, which are stripped for point clouds. We describe the reusable per-shape processing for \(\mathcal X\); the same construction is applied independently to \(\mathcal Y\). A category-specific parametric model maps \(z=(\Theta_s,\Theta_p,\Theta_a)\) to vertices \(V(z)\in\mathbb R^{M\times3}\) and joints \(J(z)\in\mathbb{R}^{K\times3}\), where \(\Theta_s,\Theta_p,\Theta_a\) are shape, pose, and auxiliary parameters. All instances share the \(M\)-vertex template \(\mathcal T=(\bar V,F_{\mathcal T})\) with $K$ joints. A \emph{canonical site} on the template therefore denotes a semantic vertex index, not a common 3D position. We use MHR~\cite{ferguson2025mhr} for humans and SMAL~\cite{zuffi20173d} for four-legged animals.

We first render \(\mathcal X\) into multi-view images with known camera parameters \((E_v,I_v)\). For rendered view \(v\), a frozen monocular estimator predicts initial parametric model parameters \(\hat z_v\) and its image-space camera parameters \((\hat E_v,\hat I_v)\). We use SAM 3D Body~\cite{yang2025sam3dbody} for humans and use PRIMA~\cite{yu2026prima} for four-legged animals.

\subsection{From Multi-View Predictions to Scan-Frame Evidence}
\label{sec:mv-grounding}

\cref{sec:prelim} provides independent view-wise pose and shape hypotheses; they might match the rendered silhouette yet be wrong in depth, scale, limb orientation, or occluded regions. We therefore ground them to the input scan before fitting a proxy. For \(\mathcal X\), this stage outputs retained views \(\mathcal V\), proposal-to-scan transforms \(\{G_v\}_{v\in\mathcal V}\), and a scan-frame skeleton \(J^{\mathrm{tri}}\) with valid joints \(\mathcal K\). These quantities provide the surface and articulation constraints used in~\cref{sec:canonization}.

\noindent\textbf{1) Generate and gate proposals.}
We sample cameras around shape $\mathcal X$, render with each known camera \((E_v,I_v)\), and apply the frozen estimator to obtain proposal \((\hat z_v,\hat E_v,\hat I_v)\). We re-render each proposal with its predicted camera and retain views whose silhouette IOU is at least \(\tau_{\mathrm{IoU}}\).

\noindent\textbf{2) Recover surface placement.}
At pixels that are foreground in both renderings, we back-project the depth from both renderings through \((E_v,I_v)\) and through \((\hat E_v,\hat I_v)\), respectively. RANSAC~\cite{fischler1981random} with Umeyama alignment~\cite{umeyama2002least} estimates proposal-to-scan \(G_v\in\operatorname{Sim}(3)\) while rejecting inconsistent depth pairs. The transformed proposal supplies surface evidence aligned to the scan-frame, i.e. the 3D frame of $\mathcal X$.

\noindent\textbf{3) Recover cross-view articulation.}
We retain predicted 2D joints but discard their monocular depths. Reprojection-filtered triangulation through the known rendering cameras yields triangulated joints \(J^{\mathrm{tri}}\) and their validity mask \(\mathcal K\). Thus, \cref{sec:canonization} uses \(G_v\) to place each proposal against the scan surface and \((J^{\mathrm{tri}},\mathcal K)\) as its shared articulation target. More details like the exact proposal equation, pixel construction, and gating procedure can be found in \Cref{sec:suppl_grounding}.

\subsection{Per-Shape Geometry-Grounded Articulation}
\label{sec:canonization}

Using the scan-frame evidence from~\cref{sec:mv-grounding}, this stage (1) fits and selects the retained proposals and (2) fuses them into a fixed-index proxy \(V\) with template connectivity \(F_{\mathcal T}\) and dispersion \(\sigma^2\). These outputs feed pairwise matching in~\cref{sec:dense-corr}.

\noindent\textbf{1) Fit and select proposals.}
A direct surface fit is local and non-convex. We therefore resolve articulation from the cross-view joints before activating the surface and skeleton constraints. At step \(t\), each retained view minimizes
\begin{equation}
    \mathcal L^{(t)}_v
    =
    \lambda_{\mathrm{kp}}\mathcal L^{\mathrm{kp}}_v
    +g(t)\!\left(
        \mathcal L^{\mathrm{surf}}_v
        +\lambda_{\mathrm{skel}}\mathcal L^{\mathrm{skel}}_v
    \right),
    \label{eq:surface-objective}
\end{equation}
where the three losses align the proposal to joints \(J^{\mathrm{tri}}\), the sampled scan surface \(X\), and skeleton bone lengths, respectively. During warm-up, \(g(t)=0\), so we optimize only \(\Theta_{p,v}\). We then rank the proposals by surface error and retain the best \(\min(N_{\mathrm{select}},|\mathcal V|)\). After the warm-up stage, with \(g(t)=1\), we refine \(\Theta_{p,v}\) and \(\Theta_{s,v}\) for the selected views using Adam~\cite{DiederikPKingma2014Arxiv}; auxiliary parameters $\Theta_{a, v}$ remain fixed. Sampling \(N_{\mathrm{samp}}\) points bounds the surface-loss cost without forming an all-vertex Chamfer matrix. Exact losses and fitted-view definitions are in~\cref{sec:suppl_optimization}.

\noindent\textbf{2) Fuse selected proposals.}
Surface and joint errors, silhouette overlap, and the RANSAC inlier ratio give each selected fit \(V_v\) a global weight \(w_v\).
\begin{align}
    V_u
    &=
    \sum_{v\in\mathcal V^{\mathrm{sel}}}
    w_vV_{v,u},
    \label{eq:weighted-fusion}\\
    \sigma_u^{2}
    &=
    \sum_{v\in\mathcal V^{\mathrm{sel}}}
    w_v\left\|V_{v,u}-V_u\right\|_2^2.
    \label{eq:vertex-uncertainty}
\end{align}
\cref{eq:weighted-fusion} fuses vertices from all retained views based on the weight. Vertex-wise dispersion is estimated by~\cref{eq:vertex-uncertainty}, i.e. the empirical cross-view disagreement. Please refer to~\cref{sec:suppl_optimization} for more details.

\subsection{Sparse Canonical Composition}
\label{sec:dense-corr}

Applying the per-shape construction to both inputs gives source and target proxies \((V_X,\sigma_X^2)\) and \((V_Y,\sigma_Y^2)\). This stage maps \(\mathcal X\) to \(\mathcal Y\) through source-to-canonical assignment, canonical-to-target assignment, and sparse composition with mode-specific filtering.

\noindent\textbf{1) Map source vertices to canonical sites.}
The barycentric coordinates of source vertex \(x_n\)'s closest proxy triangle form a nonnegative, row-stochastic row of \(A_X\in\mathbb R^{N_X\times M}\) with at most three nonzeros. Proxy distance, dispersion, and normal agreement set its relative reliability.

\noindent\textbf{2) Map canonical sites to the target.}
We construct the target-to-canonical assignment \(A_Y\) analogously. For each canonical site \(u\), we sum the contributions from associated target vertices, weighted by their local surface area and reliability, to obtain the site evidence \(e_Y(u)\). For \(e_Y(u)>0\), dividing each contribution by this sum gives soft weights over those target vertices. In full mode, these weights sum to one. In partial mode, we multiply them by the site's occupancy score and assign the remaining mass to an unmatched state \(\varnothing\). The resulting canonical-to-target assignment, including the null column, is \(Q_Y\in\mathbb R^{M\times(N_Y+1)}\).

\noindent\textbf{3) Compose and filter.}
The forward soft correspondence is
\begin{equation}
    P_{X\rightarrow Y}(n,m)
    =
    \sum_{u=1}^{M}A_X(n,u)Q_Y(u,m).
    \label{eq:soft-composition}
\end{equation}
Sparse multiplication avoids a dense \(N_X\times(N_Y+1)\) matrix. Partial mode accepts a real target only if its mass exceeds null; ties remain unmatched. For more details, please refer to \cref{sec:suppl_composition}.

\subsection{Caching, Scaling, and Optional Refinement}
\label{sec:scalability}

\noindent\textbf{Per-shape cache.}
For each shape, we cache its proxy, dispersion, sparse assignment, reliability, site evidence, and mode-specific reverse assignment; for the source these are \(V_X,\sigma_X^2,A_X,q_X,e_X,Q_X\), with analogous \(Y\)-qualified quantities for the target. For \(L\) shapes, inference and optimization run \(L\) times rather than per pair.

\noindent\textbf{Pair query.}
A query composes cached \(A_XQ_Y\). Since \(A_X\) has at most \(3N_X\) nonzeros, it avoids dense allocation; assignment and extraction remain resolution- and support-dependent (\cref{sec:suppl_scalability}).

\noindent\textbf{Optional refinement.}
The primary pipeline needs no input LBO. Spectral map refinement is optional, such as spectral ICP~\cite{ovsjanikov2012functional} and ZoomOut~\cite{melzi2019zoomout}. See~\cref{sec:implement,sec:suppl_runtime} of the Appendix for more details of settings and runtimes.

\section{Experiments}
\label{sec:experiment}

We evaluate \ourmethod under non-isometric deformation, near-isometry, high-resolution raw scans, and partial shapes, followed by component ablations. Unless stated otherwise, we report the standard target-domain mean geodesic error $\times100$ ($\downarrow$). We first average normalized target geodesic distances within each directed test pair, then average the resulting per-pair errors with equal weight across test pairs (\cref{eq:eval-geodesic}). Non-isometric and near-isometric baselines follow the UniMatch common-protocol evaluation~\cite{xiao2026universal}, except Synchronous Diffusion~\cite{cao2024synchronous}. The raw-scan table instead uses the template-domain metric in~\cref{sec:geodesic-eval}, only for within-table comparison.

\begin{table}[tb]
    \centering
    \begingroup
    \setlength{\tabcolsep}{1pt}
    \renewcommand{\arraystretch}{1.04}
    \scriptsize
    \begin{tabular}{@{}lccccc@{}}
        \toprule
        Method & Unsup. & \shortstack{No Corr.\\Tr.} & Zero-Shot & SMAL & TOPKIDS \\
        \midrule

        \multicolumn{6}{c}{\textit{Axiomatic Methods}} \\
        ZoomOut & \checkmark & \checkmark & \checkmark & 38.4 & 33.7 \\
        Smooth Shells & \checkmark & \checkmark & \checkmark & 36.1 & 11.8 \\
        DiscreteOp & \checkmark & \checkmark & \checkmark & 38.1 & 35.5 \\

        \midrule
        \multicolumn{6}{c}{\textit{Functional Map Methods}} \\
        UnsupFMNet & \checkmark & \xmark & \xmark & -   & 38.5 \\
        SURFMNet & \checkmark & \xmark & \xmark & -   & 48.6 \\
        AttentiveFMaps & \xmark & \xmark & \xmark & 5.4 & 23.4 \\
        URSSM & \checkmark & \xmark & \xmark & 6.0 & 8.9 \\
        Synchronous Diff. & \checkmark & \xmark & \xmark & \secondplace{3.6} & 5.4 \\

        \midrule
        \multicolumn{6}{c}{\textit{Semantic Methods}} \\
        Diff3F & \checkmark & \checkmark & \checkmark & 28.4 & 31.0 \\
        DenseMatcher & \xmark & \xmark & \xmark & 4.7 & 6.2 \\
        
        \midrule
        \multicolumn{6}{c}{\textit{Template-based Methods}} \\
        \ourmethod (ours) & \checkmark & \checkmark & \checkmark & \firstplace{2.4} & \secondplace{2.9} \\
        \ourmethod (+Spec.-ICP) & \checkmark & \checkmark & \checkmark & 3.7 & \firstplace{2.3} \\
        \ourmethod (+ZoomOut) & \checkmark & \checkmark & \checkmark & 6.0 & 3.3 \\
        \bottomrule
    \end{tabular}
    \endgroup
    \caption{Mean geodesic error $\times100$ ($\downarrow$) for non-isometric matching. ``No Corr.\ Tr.'' means no target-benchmark correspondence training. Baseline values follow UniMatch~\cite{xiao2026universal}, except Synchronous Diffusion~\cite{cao2024synchronous}. \tabclrscheme}
    \label{tab:exp_non_isometric}
\end{table}

\subsection{Non-Isometric Shape Matching} \label{sec:exp_non_isometric}
\noindent\textbf{Setup.} We evaluate remeshed SMAL~\cite{zuffi20173d} (animals) and TOPKIDS~\cite{lahner2016shrec} (humans). Comparisons cover axiomatic methods~\cite{melzi2019zoomout,eisenberger2020smooth,ren2021discrete}, functional-map methods~\cite{halimi2019unsupervised,roufosse2019unsupervised,li2022learning,cao2023unsupervised,cao2024synchronous}, and semantic methods~\cite{dutt2024diffusion,zhu2025densematcher}; \cref{tab:exp_non_isometric} also records their correspondence-training and zero-shot dependencies.

\begin{figure*}[t]
    \centering
    \includegraphics[width=\linewidth]{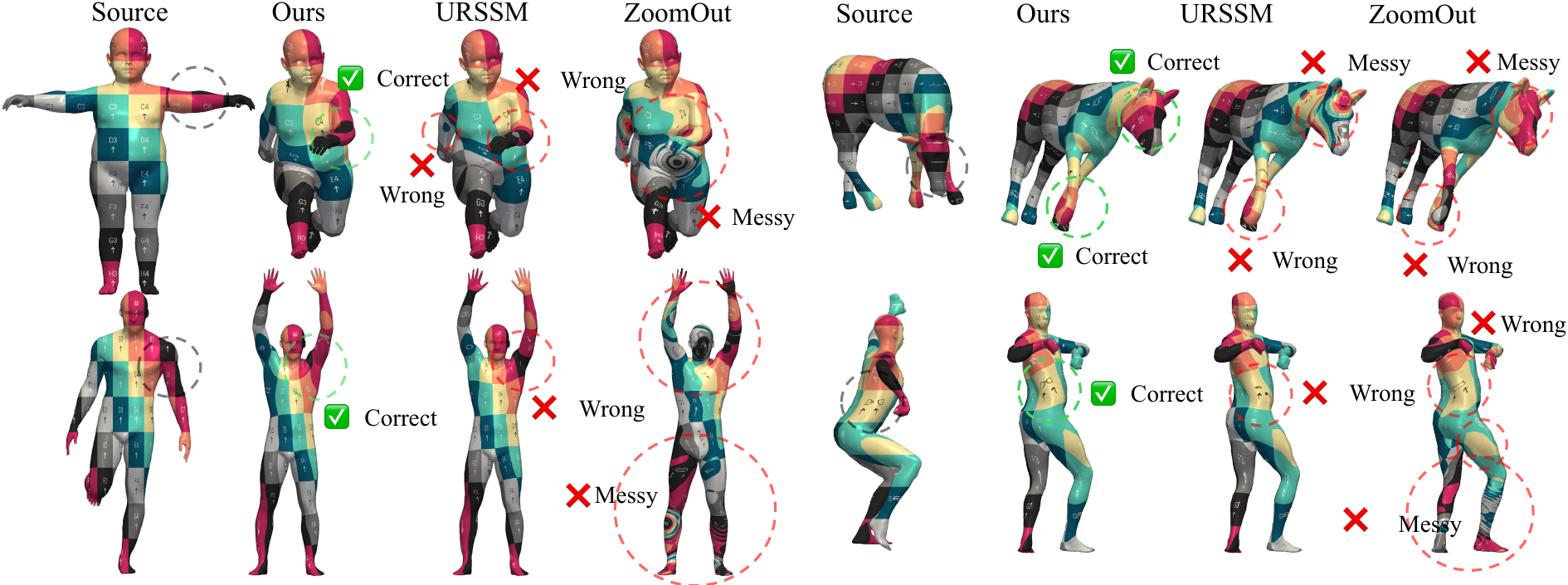}
    \caption{Selected texture-transfer examples on TOPKIDS, SMAL, FAUST, and SCAPE. The examples illustrate coherent transfers by \ourmethod and representative baseline failures.}
    \label{fig:curated}
\end{figure*}

\noindent\textbf{Results and analysis.} Before spectral refinement, \ourmethod achieves mean geodesic errors of 2.4 on SMAL and 2.9 on TOPKIDS (\cref{tab:exp_non_isometric}), 33\% and 46\% below Synchronous Diffusion and 49\% and 53\% below DenseMatcher. It also has the lowest error averaged across the two datasets before refinement. Spectral ICP improves TOPKIDS to 2.3 but worsens SMAL to 3.7, and ZoomOut refinement worsens both values. \Cref{fig:curated} provides qualitative examples.

\subsection{Near-Isometric Shape Matching}  \label{sec:exp_isometric}

\noindent\textbf{Setup.} We test whether the extrinsic \textit{articulate-then-match} paradigm remains competitive with intrinsic methods on remeshed FAUST~\cite{bogo2014faust}, SCAPE~\cite{anguelov2005scape}, and SHREC19~\cite{melzi2019shrec}. The table includes axiomatic, functional-map, semantic, and template baselines~\cite{ren2018continuous,melzi2019zoomout,eisenberger2020smooth,halimi2019unsupervised,roufosse2019unsupervised,cao2023unsupervised,cao2024synchronous,dutt2024diffusion,zhu2025densematcher,xiao2026universal,groueix20183d}.

\noindent\textbf{Results and analysis.} Before spectral refinement, \ourmethod achieves mean geodesic errors of 1.5/1.8/3.0 on FAUST/SCAPE/SHREC19 (\cref{tab:exp_isometric}). It ties Synchronous Diffusion on FAUST/SCAPE and lowers SHREC19 from 3.4 to 3.0. Adding spectral ICP yields 1.3/1.7/3.1, increasing the SHREC19 error from 3.0 to 3.1.

\begin{table}[tb]
    \centering
    \begingroup
    \setlength{\tabcolsep}{4pt}
    \renewcommand{\arraystretch}{1.04}
    \scriptsize
    \begin{tabular}{lccc}
        \toprule
        Method & FAUST & SCAPE & SHREC19 \\
        \midrule
        \multicolumn{4}{c}{\textit{Axiomatic Methods}} \\
        BCICP & 6.4 & 11.0 & 8.0 \\
        ZoomOut & 6.1 &  7.5 & 7.8 \\
        Smooth Shells & 2.5 &  4.7 & 7.6 \\
        \midrule
        \multicolumn{4}{c}{\textit{Functional Map Methods}} \\
        UnsupFMNet & 4.8 &  9.6 & 11.1 \\
        SURFMNet & 2.5 &  6.0 & 4.8 \\
        URSSM & 1.6 & 1.9 & 5.7 \\
        Synchronous Diff. & \secondplace{1.5} & \secondplace{1.8} & 3.4 \\
        \midrule
        \multicolumn{4}{c}{\textit{Semantic Methods}} \\
        Diff3F & 20.7 & 22.1 & 26.3 \\
        DenseMatcher & 1.6 & 2.0 & \secondplace{3.1} \\
        UniMatch & 1.6 & 1.9 & 3.2 \\
        \midrule
        \multicolumn{4}{c}{\textit{Template-based Methods}} \\
        3D-CODED & 2.5 & 9.8 & 7.7 \\
        \ourmethod (ours) & \secondplace{1.5}  & \secondplace{1.8} & \firstplace{3.0} \\
        \ourmethod (+Spec.-ICP) & \firstplace{1.3}  & \firstplace{1.7} & \secondplace{3.1} \\
        \ourmethod (+ZoomOut) & \secondplace{1.5}  & 1.9 & 3.2 \\
        \bottomrule
    \end{tabular}
    \endgroup
    \caption{Mean geodesic error $\times100$ ($\downarrow$) for near-isometric matching. Baselines follow UniMatch~\cite{xiao2026universal}; the Synchronous Diffusion row follows~\cite{cao2024synchronous}. \tabclrschemefollow}
    \label{tab:exp_isometric}
\end{table}

\begin{figure}[tb]
    \centering
    \includegraphics[width=0.9\linewidth]{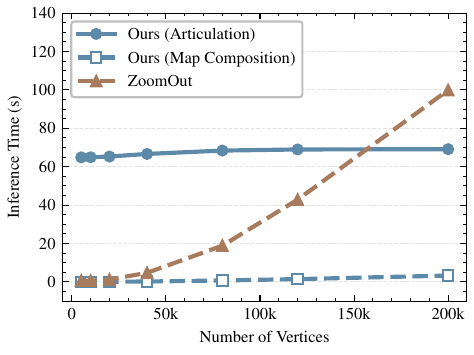}
    \caption{Runtime decomposition across resolution. Articulation is measured in seconds per shape; \ourmethod composition and chunked ZoomOut are measured in seconds per directed pair. Articulation varies from about 65 to 69 seconds per shape, composition grows to 3.33 seconds per pair, and ZoomOut reaches about 100 seconds per pair at raw resolution.}
    \label{fig:scalability}
\end{figure}

\subsection{Scalability to High-Resolution Raw Scans} \label{sec:exp_wild_scans}

\noindent\textbf{Setup.} We decimate all 100 raw FAUST scans~\cite{bogo2014faust} from 160k--200k vertices to 5k--120k. At every resolution, quantitative evaluation uses the same 400 shape pairs over the 20-shape test split and the same template-domain metric and ground truth (\cref{sec:geodesic-eval}); its values are not interchangeable with~\cref{eq:eval-geodesic}. ZoomOut uses the chunked nearest-search implementation in~\cref{sec:suppl_runtime}.

\noindent\textbf{Results and analysis.} \ourmethod decreases from 2.42 at 5k to 2.11 at raw resolution and stays below 2.5 (\cref{tab:exp_wild_scans}). Chunked ZoomOut remains at 19.89--20.92; the evaluated URSSM configurations span 10.30--20.80 through 20k and report implementation-specific OOM from 40k. \Cref{fig:scalability} shows nearly stable per-shape articulation but growing per-pair composition runtime, see Appendix~\ref{sec:suppl_runtime} for more details. \Cref{fig:raw} provides qualitative examples.

\begin{table}[tb]
    \centering
    \begingroup
    \setlength{\tabcolsep}{1.8pt}
    \renewcommand{\arraystretch}{1.04}
    \scriptsize
    \begin{tabular}{lccccccc}
        \toprule
        Method & 5k & 10k & 20k & 40k & 80k & 120k & Raw \\
        \midrule
        ZoomOut (ch.) & 20.92 & 20.54 & 20.24 & \secondplace{20.57} & \secondplace{20.31} & \secondplace{20.21} & \secondplace{19.89} \\
        URSSM~(xyz) & 14.30 & \secondplace{13.70} & \secondplace{11.60} & OOM & OOM & OOM & OOM \\
        URSSM~(wks) & \secondplace{10.30} & 20.80 & 15.90 & OOM & OOM & OOM & OOM \\
        \ourmethod~(ours) & \firstplace{2.42} & \firstplace{2.25} & \firstplace{2.18} & \firstplace{2.16} & \firstplace{2.16} & \firstplace{2.13} & \firstplace{2.11} \\
        \bottomrule
    \end{tabular}
    \endgroup
    \caption{Template-domain mean geodesic error $\times100$ ($\downarrow$) on 400 directed raw-FAUST pair queries. ZoomOut uses chunked nearest search. OOM denotes out of memory error.}
    \label{tab:exp_wild_scans}
\end{table}

\begin{figure*}[t]
    \centering
    \includegraphics[width=0.9\linewidth]{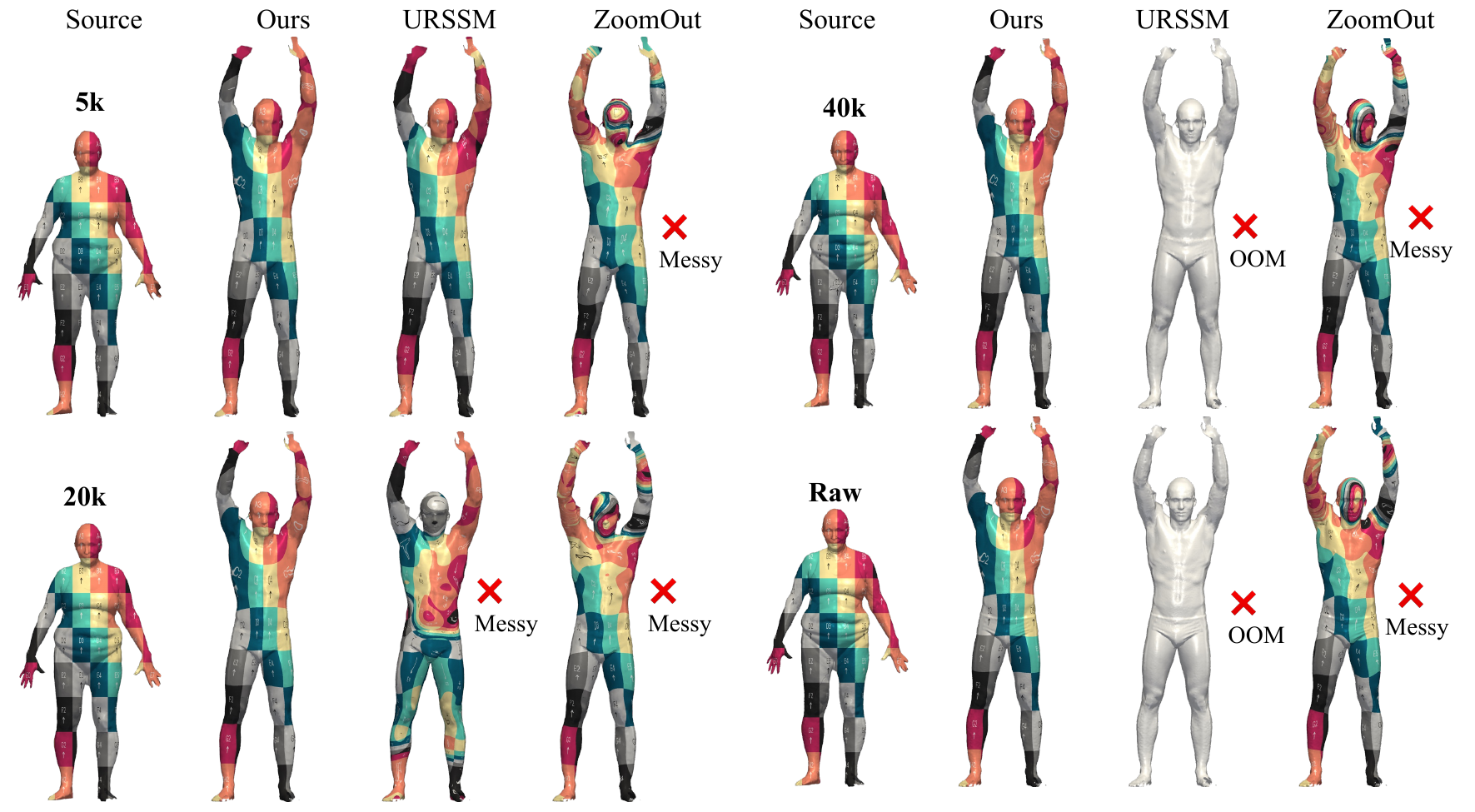}
    \caption{Texture transfer on original FAUST raw scans. 5k, 20k, 40k, and original 160k--200k-vertex scans are shown.}
    \label{fig:raw}
\end{figure*}

\subsection{Partial Shape Matching}
\label{sec:exp_partial_scans}

\noindent\textbf{Setup.} We use the BeCoS processing pipeline~\cite{ehm2025beyond} independently on FAUST~\cite{bogo2014faust} and KIDS~\cite{rodola2014dense}, applying random rotation, remeshing, and ray-cropping to simulate partial laser scans. We evaluate partial-to-full (P-F) and partial-to-partial (P-P) matching against ZoomOut~\cite{melzi2019zoomout} and two URSSM~\cite{cao2023unsupervised} input configurations. URSSM is trained from scratch for 30 epochs on each corresponding training split.

\noindent\textbf{Results and analysis.} \ourmethod achieves the lowest mean geodesic error among the evaluated methods in all four settings: 2.7/3.3 on FAUST P-F/P-P and 6.7/5.4 on KIDS (\cref{tab:exp_partial_scans}). Qualitative examples appear in~\cref{fig:partial_scans}.

\begin{table}[tb]
    \centering
    \begingroup
    \setlength{\tabcolsep}{7pt}
    \renewcommand{\arraystretch}{1.04}
    \small
    \begin{tabular}{lcccc}
        \toprule
        & \multicolumn{2}{c}{FAUST} & \multicolumn{2}{c}{KIDS} \\
        \cmidrule(lr){2-3} \cmidrule(lr){4-5}
        \multicolumn{1}{c}{Method} & P-F & P-P & P-F & P-P \\
        \midrule
        ZoomOut & 42.2 & 68.7 & \secondplace{30.0} & \secondplace{40.1} \\
        URSSM(xyz) & \secondplace{22.4} & \secondplace{20.6} & 40.2 & 41.4 \\
        URSSM(wks) & 31.9 & 27.3 & 42.7 & 47.6 \\
        \ourmethod (ours) & \firstplace{2.7} & \firstplace{3.3} & \firstplace{6.7} & \firstplace{5.4} \\
        \bottomrule
    \end{tabular}
    \endgroup
    \caption{Mean geodesic error $\times100$ ($\downarrow$) for BeCoS-processed FAUST and KIDS subsets. P-F and P-P denote partial-to-full and partial-to-partial matching. The comparison reports correspondence error; the fraction of queries receiving a match is not reported.}
    \label{tab:exp_partial_scans}
\end{table}

\subsection{Ablation Study} \label{sec:exp_ablation}

\noindent\textbf{Setup.} On SMAL~\cite{zuffi20173d}, TOPKIDS~\cite{lahner2016shrec}, and FAUST~\cite{bogo2014faust}, we compare the full pipeline with: (1) ungrounded averaging of multi-view predictions, (2) no keypoint-based pose stage, and (3) no surface-refinement stage. Every articulation variant uses both evidence-weighted soft composition and nearest hard composition.

\noindent\textbf{Results and analysis.} Full soft composition is lowest on all three datasets at 2.4/2.9/1.5 (\cref{tab:exp_ablation}). Removing grounding causes the largest degradation (6.5/6.1/2.6), and omitting either stage worsens every dataset. Soft composition yields lower error in every paired comparison; for the full pipeline, switching from hard to soft composition reduces error from 3.7 to 2.4, 3.0 to 2.9, and 1.6 to 1.5.

\begin{table}[!b]
    \centering
    \begingroup
    \setlength{\tabcolsep}{7pt}
    \renewcommand{\arraystretch}{1.04}
    \small
    \begin{tabular}{lccc}
        \toprule
        Variant & SMAL & TOPKIDS & FAUST \\
        \midrule
        \multicolumn{4}{c}{\textit{Evidence-weighted soft map composition}} \\
        w/o grounding      & 6.5 & 6.1 & 2.6 \\
        w/o keypoint stage & 3.0 & 4.3 & 1.7 \\
        w/o surface stage  & 5.6 & 3.9 & 1.9 \\
        full               & \firstplace{2.4} & \firstplace{2.9} & \firstplace{1.5} \\
        \midrule
        \multicolumn{4}{c}{\textit{Nearest hard composition}} \\
        w/o grounding      & 6.8 & 6.2 & 2.7 \\
        w/o keypoint stage & 3.1 & 4.4 & 1.8 \\
        w/o surface stage  & 6.0 & 4.0 & 2.0 \\
        full               & 3.7 & 3.0 & 1.6 \\
        \bottomrule
    \end{tabular}
    \endgroup
    \caption{Mean geodesic error $\times100$ ($\downarrow$) for articulation ablations with evidence-weighted soft and nearest hard composition. Values are aggregate point estimates.}
    \label{tab:exp_ablation}
\end{table}

\section{Conclusion}
\label{sec:conclusion}

We presented \ourmethod, a zero-shot framework for accurate and scalable dense correspondence on in-the-wild raw meshes. Geometry-Grounded Articulation uses each scan's geometry to refine and fuse multi-view template predictions into a proxy with consistent vertex indices across shapes. We cache the proxy and its sparse assignments for reuse across shape pairs. Sparse composition then produces correspondences without repeating per-shape inference or optimization. The matching pipeline requires neither target-dataset correspondence training nor LBO computation on the input meshes. On raw FAUST and its decimated versions, the template-domain mean geodesic error ($\times100$) stays below 2.5 across resolutions from 5k to 160k--200k vertices, demonstrating stable matching accuracy as mesh resolution increases. Experiments on non-isometric and partial shapes also show low correspondence errors. Together, these results demonstrate the value of grounding articulation once per scan and reusing it for scalable matching of raw shape collections.

\newpage
\bibliography{main}

\clearpage
\appendix
\setcounter{section}{0}
\setcounter{figure}{0}
\renewcommand{\thefigure}{S\arabic{figure}}
\setcounter{table}{0}
\renewcommand{\thetable}{S\arabic{table}}
\setcounter{page}{1}

\addtocontents{toc}{\protect\setcounter{tocdepth}{2}}

\twocolumn[{%
\begin{center}
\LARGE \textbf{Geometry-Grounded Articulation and Test-Time Adaptation for Zero-Shot, Scalable Shape Matching in the Wild}\\[0.6em]
\LARGE \textit{Supplementary Material}
\end{center}

}]

This supplementary material expands the method derivation in~\cref{sec:suppl_method_details}, provides implementation details in~\cref{sec:implement}, reports robustness experiments in~\cref{sec:suppl_robustness}, clarifies animal benchmarks and the use of the trained prior in~\cref{sec:suppl_animals}, provides details of the uncurated high-resolution mesh benchmarks in~\cref{sec:suppl_runtime}, and describes the memory-efficient geodesic evaluation used for the high-resolution uncurated-mesh experiment in~\cref{sec:geodesic-eval}.

\section{Additional Method Details}
\label{sec:suppl_method_details}

This section expands the notation, geometric construction, objectives, and map-extraction details summarized in~\cref{sec:methodology}. No additional model training or fine-tuning stage is introduced.

\begin{table*}[t]
    \centering
    \footnotesize
    \setlength{\tabcolsep}{4pt}
    \renewcommand{\arraystretch}{1.08}
    \begin{tabular}{@{}p{0.14\textwidth}p{0.32\textwidth}p{0.14\textwidth}p{0.32\textwidth}@{}}
        \toprule
        Symbol & Meaning & Symbol & Meaning \\
        \midrule
        \(\mathcal X,\mathcal Y\) &
        Source and target meshes, \(\mathcal X=(X,F_X)\) and \(\mathcal Y=(Y,F_Y)\) &
        \(n,m;\ u,v\) &
        Source/target raw-vertex indices; canonical-site and rendering-view indices \\

        \(X,Y\) &
        Source and target raw-vertex matrices \(X=[x_n]\) and \(Y=[y_m]\) &
        \(\mathcal T=(\bar V,F_{\mathcal T})\) &
        Shared parametric template with \(M\) canonical sites \\

        \(z=(\Theta_s,\Theta_p,\Theta_a)\) &
        Shape, pose, and auxiliary model parameters &
        \(V(z),J(z)\) &
        Parametric-model vertices and joints \\

        \((E_v,I_v)\) &
        Known camera used to render the current shape &
        \((\hat E_v,\hat I_v)\) &
        Camera predicted by the monocular estimator \\

        \(\mathcal V,\mathcal V^{\mathrm{sel}}\) &
        Retained views and the selected subset used for fusion &
        \(G_v\in\operatorname{Sim}(3)\) &
        Similarity transform from proposal to scan coordinates \\

        \(J^{\mathrm{tri}},\mathcal K\) &
        Triangulated scan-frame skeleton and valid joint set &
        \(V_v,V\) &
        Fitted template for view \(v\) and fused proxy \\

        \(r_v,w_v\) &
        View-quality residual and normalized fusion weight &
        \(\sigma^2\) &
        Per-site empirical cross-view dispersion \\

        \(A_X,A_Y\) &
        Sparse raw-to-proxy assignments for source and target &
        \(q_X,q_Y\) &
        Source and target relative reliability \\

        \(e_Y,o_Y\) &
        Target canonical-site evidence and occupancy &
        \(\begin{gathered}
            Q_Y\in\\[-2pt]
            \mathbb R^{M\times(N_Y+1)}
        \end{gathered}\) &
        Canonical-to-target assignment including the null state \\

        \(P_{X\rightarrow Y}=A_XQ_Y\) &
        Soft source-to-target correspondence composition &
        \(\hat\pi_{X\to Y},\pi_{X\to Y},\varnothing\) &
        Best real candidate, accepted hard map, and unmatched state \\

        \(s_{nm}^{\mathrm{rec}}\) &
        Symmetric reciprocal-consistency score &
        \(T_{\mathrm{pose}},g(t)\) &
        Phase boundary and continuation gate for optimization \\

        \(\begin{gathered}
            N_{\mathrm{view}},N_{\mathrm{select}},\\[-2pt]
            N_{\mathrm{samp}}
        \end{gathered}\) &
        Numbers of rendered views, selected views, and Chamfer samples &
        & \\
        \bottomrule
    \end{tabular}
    \caption{Summary of the notation used in the main Method and Appendix~A. Unqualified symbols describe the reusable per-shape processing of \(\mathcal X\); the same construction applies to \(\mathcal Y\). Shape qualifiers \(X,Y\) appear only when the pair must be distinguished. Proxies remain in their respective scan coordinate frames, and a canonical site is a shared semantic template index.}
    \label{tab:notation}
\end{table*}

\subsection{Parametric Proposals and Geometric Grounding}
\label{sec:suppl_grounding}

\noindent\textbf{Parametric model and monocular proposals.}
For shape, pose, and auxiliary parameters \(z=(\Theta_s,\Theta_p,\Theta_a)\), the category-specific model is
\begin{equation}
    \begin{aligned}
        \bigl(V(z),J(z)\bigr)
        &=
        \bigl(\mathcal M_V(z),\mathcal M_J(z)\bigr),\\
        V(z)&\in\mathbb R^{M\times3},\qquad
        J(z)\in\mathbb R^{N_J\times3}.
    \end{aligned}
    \label{eq:parametric-model}
\end{equation}
Every instance is generated from the same \(M\)-vertex template \(\mathcal T=(\bar V,F_{\mathcal T})\), so vertex \(u\) has a fixed semantic identity across pose and shape.

For known rendering camera \((E_v,I_v)\), renderer \(\mathcal R\), and frozen estimator \(\mathcal E\), the proposal from view \(v\) is
\begin{equation}
    \hat z_v,\hat E_v,\hat I_v
    =
    \mathcal E\!\left(
        \mathcal R(\mathcal X;E_v,I_v)
    \right).
    \label{eq:2dvfm}
\end{equation}
The known camera defines the rendered scan image and its rays in the scan frame; the predicted camera places the estimator's local template in that image.

\begin{figure}[tb]
    \centering
    \begin{subfigure}{0.47\linewidth}
        \centering
        \includegraphics[height=2.1cm]{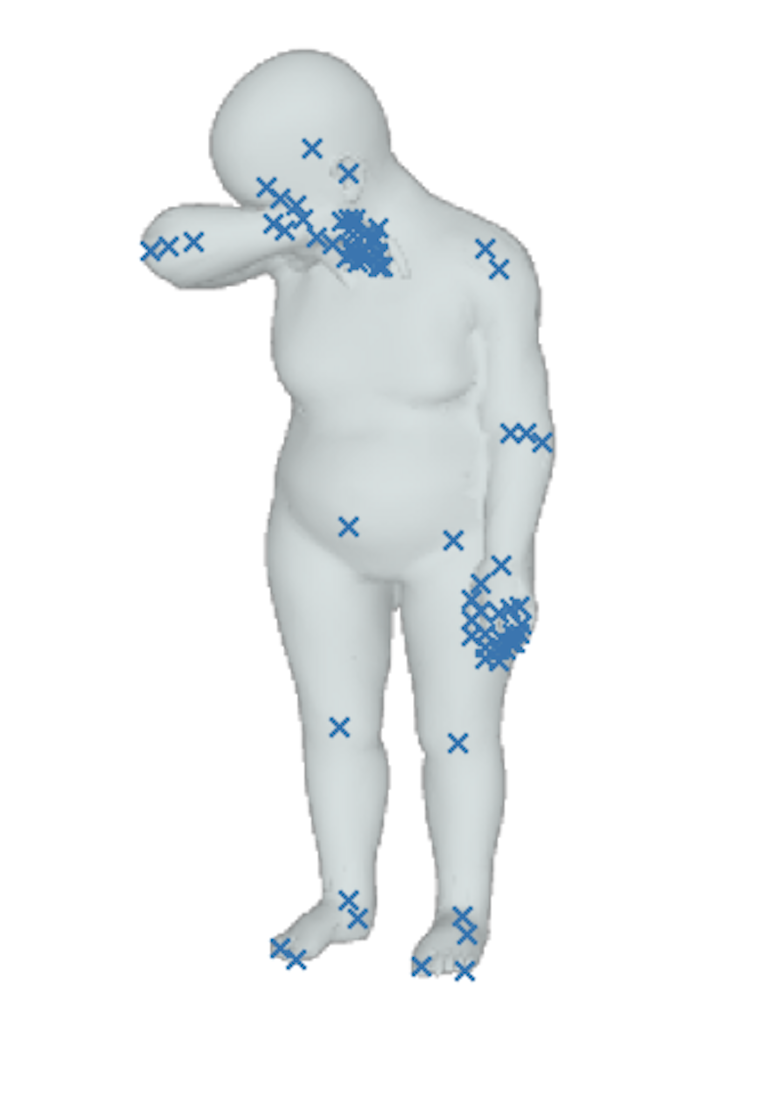}
        \hskip-1.5em
        \includegraphics[height=2.1cm]{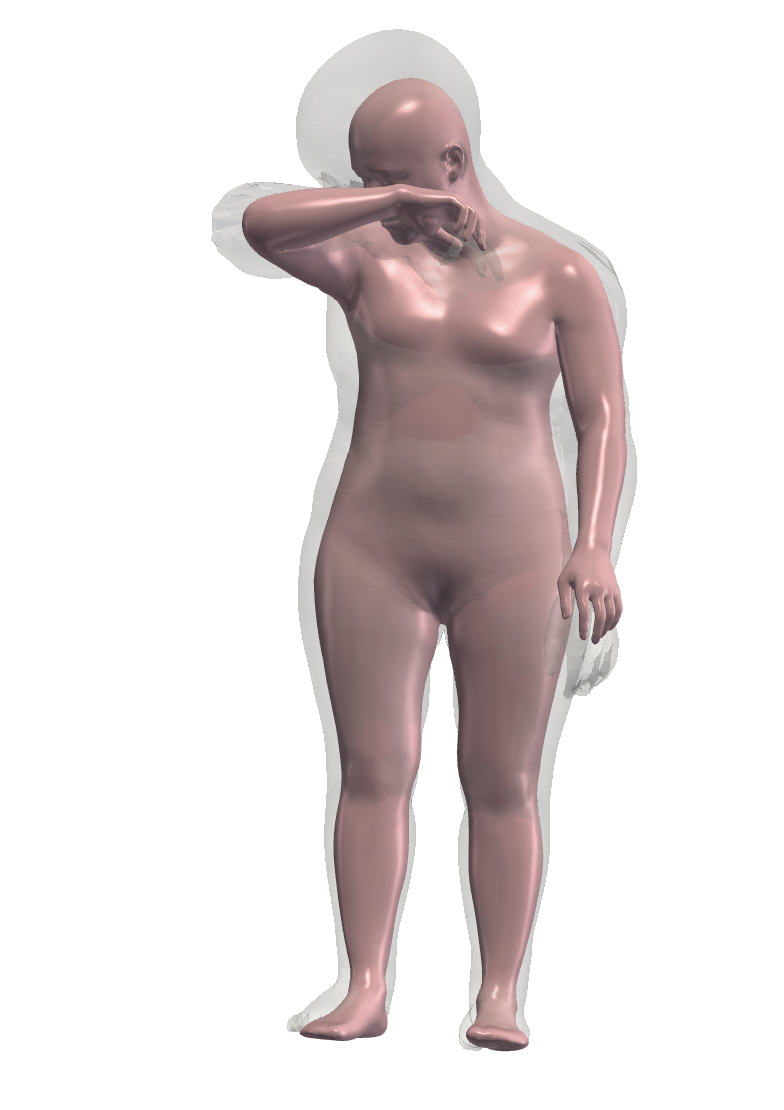}
        \hskip-1.0em
        \includegraphics[height=2.1cm]{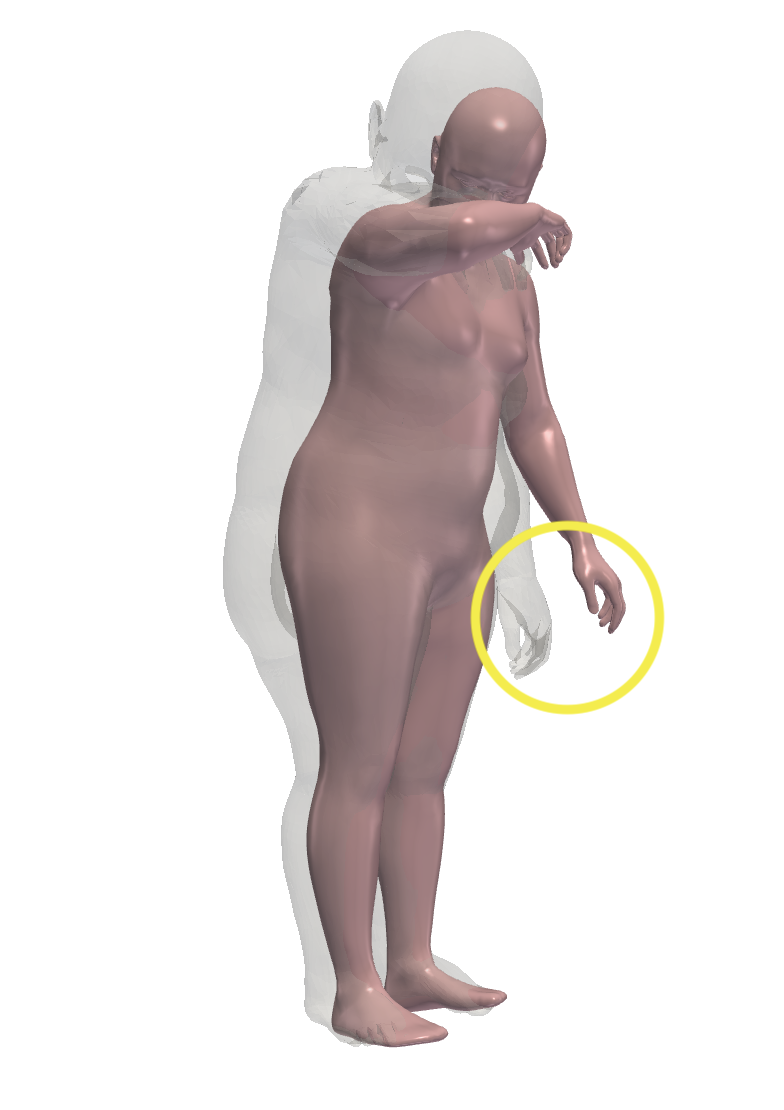}
        \caption{}
        \label{fig:2d-to-3d-ambiguity}
    \end{subfigure}
    \begin{subfigure}{0.47\linewidth}
        \centering
        \includegraphics[height=2.1cm]{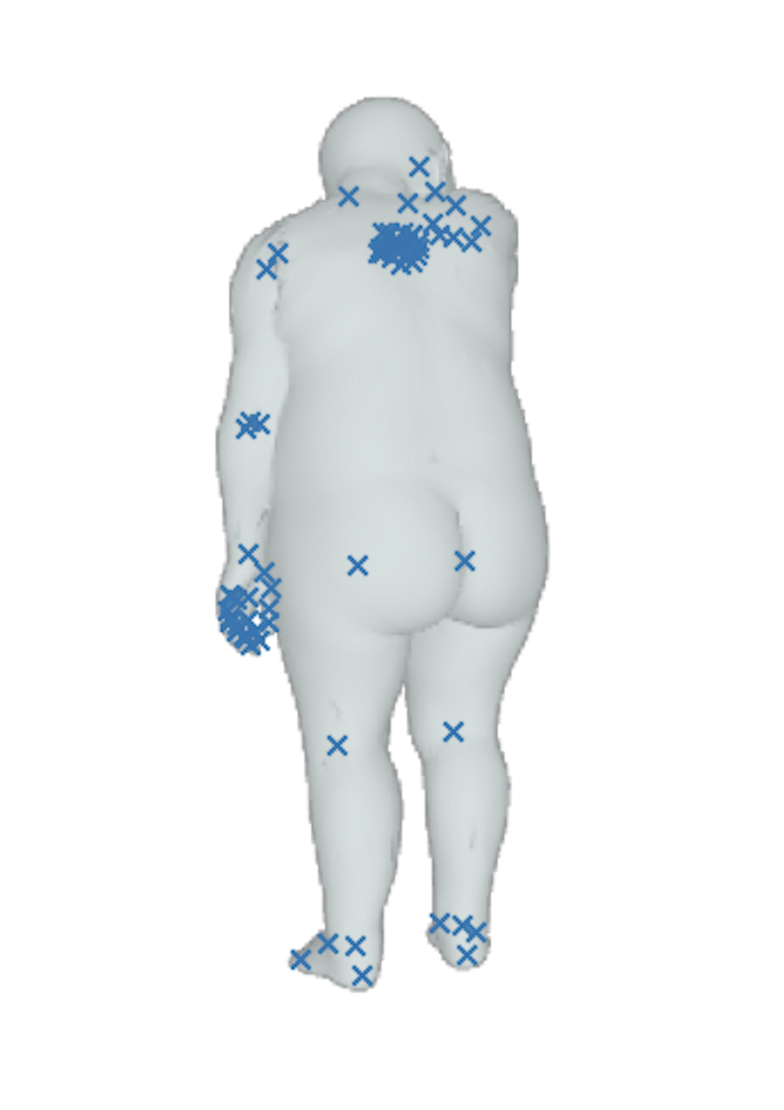}
        \hskip-1.5em
        \includegraphics[height=2.1cm]{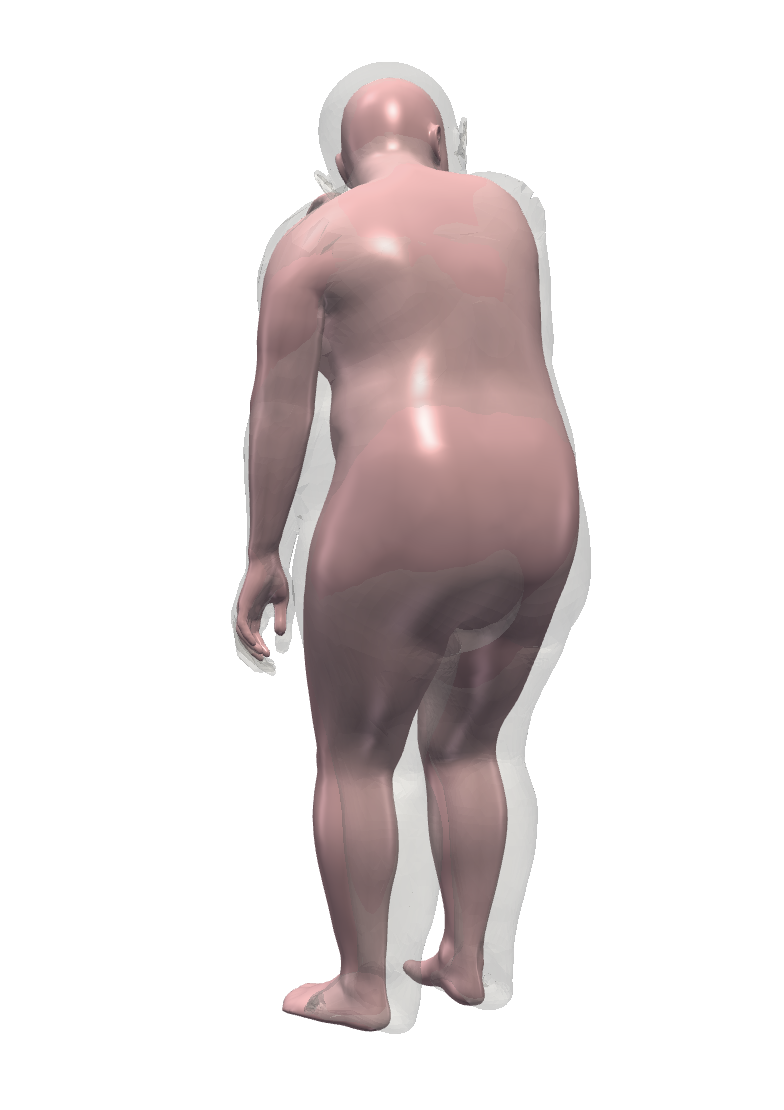}
        \hskip-1.0em
        \includegraphics[height=2.1cm]{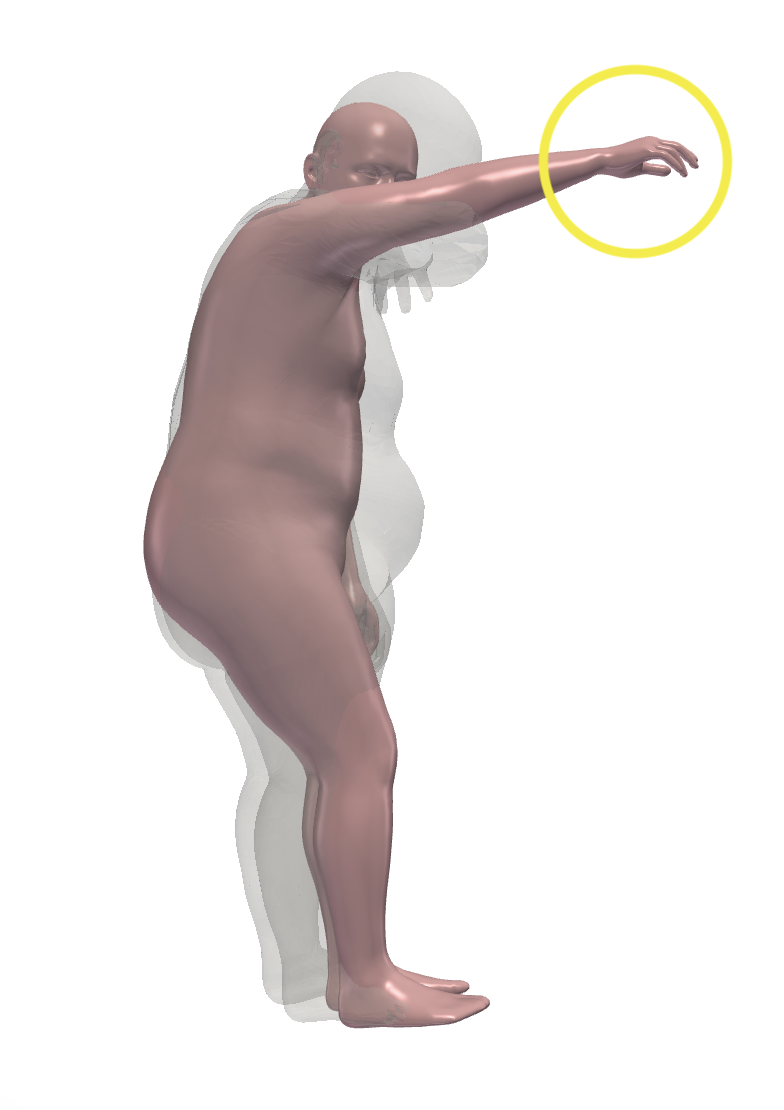}
        \caption{}
        \label{fig:partial-observation}
    \end{subfigure}
    \caption{Why multi-view grounding is necessary. (a) Accurate 2D alignment does not guarantee depth, limb orientation, or global scale alignment. (b) Self-occluded regions are prior-driven and can be mutually inconsistent across views. That said, with proper error-gating, we can leverage the keypoints projected to 2D as cross-view consistent evidence to resolve the 3D ambiguity and inconsistency issues.}
    \label{fig:why-grounding}
\end{figure}

\noindent\textbf{Rendering, view gating, and similarity alignment.}
We sample cameras around the centered input and render \(\{(E_v,I_v)\}_{v=1}^{N_{\mathrm{view}}}\). Applying~\cref{eq:2dvfm} gives one parametric proposal per view. We render each proposal with \((\hat E_v,\hat I_v)\) into the original image grid and compare its silhouette with the scan rendering. Views whose overlap \(\operatorname{IoU}_v\) falls below \(\tau_{\mathrm{IoU}}\) are discarded; the retained set is \(\mathcal V\).

At pixels that are foreground in both renderings, scan depth is back-projected through \((E_v,I_v)\), while proposal depth is back-projected through \((\hat E_v,\hat I_v)\). The resulting co-located pairs \(P^{\mathrm{in}}_v\) and \(P^{\mathrm{pred}}_v\) lie in the scan and proposal frames, respectively. We estimate \(G_v\in\operatorname{Sim}(3)\), directed from proposal to scan, using Umeyama alignment~\cite{umeyama2002least} inside RANSAC~\cite{fischler1981random}. RANSAC rejects pairs caused by occlusion, segmentation error, or inaccurate predicted geometry and returns an inlier ratio \(\eta_v\).

\noindent\textbf{Error-gated triangulation.}
We project the predicted joints with \((\hat E_v,\hat I_v)\), retain their 2D locations in the original rendering, and discard their predicted depths. Using the known rendering cameras \((E_v,I_v)\), we triangulate each joint across views, remove observations with large reprojection residuals, and triangulate again. This yields a scan-frame skeleton \(J^{\mathrm{tri}}\in\mathbb R^{N_J\times3}\) and valid joint set \(\mathcal K\). The formulation requires \(\mathcal V\ne\emptyset\) and \(\mathcal K\ne\emptyset\); extreme truncation can violate these conditions. The transforms \(G_v\) provide surface placement, while \(J^{\mathrm{tri}}\) constrains articulation.

\subsection{Optimization and Multi-View Consensus}
\label{sec:suppl_optimization}

\noindent\textbf{Continuation schedule and losses.}
The continuation weight in~\cref{eq:surface-objective} is
\begin{equation}
    g(t)
    =
    \begin{cases}
        0, & t<T_{\mathrm{pose}},\\
        1, & t\ge T_{\mathrm{pose}}.
    \end{cases}
    \label{eq:surface-curriculum}
\end{equation}
Its constituent losses are
\begin{align}
    \mathcal L^{\mathrm{kp}}_v
    &=
    \frac{1}{|\mathcal K|}
    \sum_{k\in\mathcal K}
    \left\|
        J^{\mathrm{tri}}_k
        -\bigl(G_v[\mathcal M_J(z_v)]\bigr)_k
    \right\|_2^2,
    \label{eq:keypoint-loss}\\
    \mathcal L^{\mathrm{surf}}_v
    &=
    \operatorname{SquaredChamfer}_{N_{\mathrm{samp}}}
    \!\left(
        G_v[\mathcal M_V(z_v)],X
    \right),
    \label{eq:surface-loss}\\
    \mathcal L^{\mathrm{skel}}_v
    &=
    \left\|
        L_{\mathrm{bone}}\!\left(G_v[\mathcal M_J(z_v)]\right)
        -L_{\mathrm{bone}}(J^{\mathrm{tri}})
    \right\|_2^2.
    \label{eq:skeleton-loss}
\end{align}
During the warm-up, only \(\mathcal L^{\mathrm{kp}}_v\) is active. We optimize \(\Theta_{p,v}\) while holding shape and auxiliary parameters fixed, rank the hypotheses by \(\mathcal L^{\mathrm{surf}}_v\), and retain the best \(\min(N_{\mathrm{select}},|\mathcal V|)\) views in \(\mathcal V^{\mathrm{sel}}\). In the second phase, we jointly refine \(\Theta_{p,v}\) and \(\Theta_{s,v}\), while \(\Theta_{a,v}\) remains fixed. The fitted template from view \(v\) is
\begin{equation}
    V_v=G_v[\mathcal M_V(z_v^{*})],
    \qquad v\in\mathcal V^{\mathrm{sel}}.
    \label{eq:view-fit}
\end{equation}

\noindent\textbf{View confidence.}
We summarize the global quality of view \(v\) by
\begin{equation}
    r_v
    =
    \alpha\widetilde E^{\mathrm{surf}}_v
    +\beta\widetilde E^{\mathrm{kp}}_v
    +\gamma\bigl(1-\operatorname{IoU}_v\bigr)
    +\delta\bigl(1-\eta_v\bigr),
    \label{eq:view-residual}
\end{equation}
where the surface and keypoint residuals are normalized by the squared input bounding-box diagonal and \(\alpha,\beta,\gamma,\delta\ge0\). The normalized confidence weights are
\begin{equation}
    w_v
    =
    \frac{\exp(-r_v/\tau_w)}
    {\sum_{v'\in\mathcal V^{\mathrm{sel}}}\exp(-r_{v'}/\tau_w)},
    \qquad
    v\in\mathcal V^{\mathrm{sel}},
    \label{eq:view-weight}
\end{equation}
with temperature \(\tau_w>0\). Views with non-finite residuals receive zero weight; if no finite residual remains, we use uniform weights over \(\mathcal V^{\mathrm{sel}}\).

\noindent\textbf{Weighted least-squares interpretation.}
Each fitted template supplies an observation
\begin{equation}
    V_{v,u}=V_u+\epsilon_{v,u},
    \label{eq:latent-proxy-observation}
\end{equation}
where \(\epsilon_{v,u}\in\mathbb R^3\) collects residual prediction, alignment, and fitting error. Because the observations share the same vertex ordering, the proxy in~\cref{eq:weighted-fusion} is the closed-form solution of
\begin{equation}
    V
    =
    \arg\min_{\widetilde V\in\mathbb R^{M\times3}}
    \sum_{v\in\mathcal V^{\mathrm{sel}}}
    w_v\|\widetilde V-V_v\|_F^2.
    \label{eq:weighted-fusion-wls}
\end{equation}
The per-view weights do not capture spatially varying disagreement. Vertex-wise dispersion is estimated in~\cref{eq:vertex-uncertainty}. The fused proxy remains in scan coordinates and need not lie exactly on the parametric-model manifold.

\subsection{Sparse Assignment and Map Extraction}
\label{sec:suppl_composition}

\noindent\textbf{Sparse barycentric canonicalization.}
Applying the preceding per-shape construction to \(\mathcal X\) and \(\mathcal Y\) yields proxies \((V_X,\sigma_X^2)\) and \((V_Y,\sigma_Y^2)\), respectively. For source vertex \(x_n\), let its closest point on \(V_X\) lie on triangle \((u_1,u_2,u_3)\) with barycentric coordinates \((b_1,b_2,b_3)\). We define
\begin{equation}
    A_X(n,u_k)=b_k,
    \qquad k\in\{1,2,3\},
    \label{eq:barycentric-A}
\end{equation}
with all other entries zero. Thus \(A_X\in\mathbb R^{N_X\times M}\) is nonnegative and row-stochastic, with at most three nonzeros per row. Its interpolated empirical dispersion is
\begin{equation}
    \sigma_X^2(x_n)=A_X(n,:)\sigma_X^2.
    \label{eq:interpolated-uncertainty}
\end{equation}
Let \(\nu_n^X\) and \(\hat\nu_n^X\) be the raw and proxy normals at that correspondence. The relative reliability is
\begin{equation}
    \begin{aligned}
        q_X(x_n)
        ={}&
        \exp\!\left(-\frac{d_X(x_n)^2}{2s_d^2}\right)
        \exp\!\left(-\frac{\sigma_X^2(x_n)}{2s_u^2}\right)\\
        &{}\cdot\max\!\left(0,(\nu_n^X)^{\top}\hat\nu_n^X\right),
    \end{aligned}
    \label{eq:raw-confidence}
\end{equation}
where \(d_X(x_n)\) is the distance to the closest proxy point. \(s_d\) and \(s_u\) are scaled by the proxy bounding-box diagonal.

\noindent\textbf{Area-corrected target evidence and occupancy.}
Irregular tessellation can place many vertices in a small surface region. We use lumped target area
\(\omega_m^Y=\frac{1}{3}\sum_{f\ni y_m}\operatorname{area}(\text{face})\) to scale target evidence (points $y_m$ on the target shape that is associated with canonical site $u$, in a reverse map $A(m, u)$)
\begin{equation}
    e_Y(u)
    =
    \sum_{m=1}^{N_Y}
    \omega_m^Yq_Y(y_m)A_Y(m,u).
    \label{eq:target-evidence}
\end{equation}
For full matching, every site is assumed to have associated target vertices. For partial matching, evidence becomes a soft occupancy ratio:
\begin{equation}
    \begin{aligned}
        o_Y(u)
        &=
        \begin{cases}
            1, & \text{full matching},\\[2pt]
            1-\exp\!\left(-e_Y(u)/\kappa_Y\right),
            & \text{partial matching},
        \end{cases}\\
        \kappa_Y
        &=
        \rho\,\frac{\operatorname{area}(\mathcal Y)}{M}.
    \end{aligned}
    \label{eq:occupancy}
\end{equation}
Here \(\rho\) is dimensionless. Low evidence marks a potentially absent 3D target region; \(o_Y\) is neither a calibrated overlap probability nor 2D visibility.

\noindent\textbf{Canonical-to-target reverse assignment.}
The last column of \(Q_Y\in\mathbb R^{M\times(N_Y+1)}\) represents \(\varnothing\). For a real target vertex \(m\le N_Y\),
\begin{equation}
Q_Y(u,m)
=
\begin{cases}
o_Y(u)
\dfrac{\omega_m^Y q_Y(y_m) A_Y(m,u)}
{\displaystyle\sum_{j=1}^{N_Y}
\omega_j^Y q_Y(y_j) A_Y(j,u) + \epsilon},
& m \neq \varnothing\\
1-o_Y(u),
& m = \varnothing.
\end{cases}
\label{eq:reverse}
\end{equation}
For \(e_Y(u)>0\), the row sums to one after including the null state. Since \(q_Y\) is relative rather than probabilistic, \(Q_Y\) is interpreted only as a soft assignment. In full mode the null column is zero; a zero-evidence row is handled during hard extraction.

\noindent\textbf{Hard-map extraction.}
After computing~\cref{eq:soft-composition}, the best real target is
\begin{equation}
    \hat\pi_{X\rightarrow Y}(n)
    =
    \arg\max_{1\le m\le N_Y}P_{X\rightarrow Y}(n,m).
    \label{eq:hard-forward}
\end{equation}
For partial matching, let
\(p_n^{+}=P_{X\to Y}(n,\hat\pi_{X\to Y}(n))\) and
\(p_n^{\varnothing}=P_{X\to Y}(n,\varnothing)\). We set
\begin{equation}
    \pi_{X\to Y}(n)
    =
    \begin{cases}
        \hat\pi_{X\to Y}(n),
        & p_n^{+}>p_n^{\varnothing},\\[2pt]
        \varnothing, & \text{otherwise}.
    \end{cases}
    \label{eq:partial-null-test}
\end{equation}
For full matching, if a vertex has zero evidence, we fall back to nearest-neighbor search using the canonical proxies as intermediates:
\begin{align}
    u^*(n)
    &=
    \arg\min_u\|x_n-V_{X,u}\|_2,
    \label{eq:fallback-canonical}\\
    \pi^{\mathrm{topo}}_{X\rightarrow Y}(n)
    &=
    \arg\min_m\|y_m-V_{Y,u^*(n)}\|_2.
    \label{eq:fallback-target}
\end{align}

\subsection{Sparse Storage and Scaling}
\label{sec:suppl_scalability}

We store \(A_X\), \(Q_Y\), and the pairwise result in compressed sparse form. The sparse multiplication computation is bounded by
\begin{equation}
    \sum_n\sum_{u\in\mathcal U_X(n)}
    \operatorname{nnz}\!\left(Q_Y(u,:)\right),
    \label{eq:sparse-work}
\end{equation}
rather than \(N_XN_Y\). The bound can still grow when many target vertices concentrate on the same canonical sites.

\section{Implementation Details} \label{sec:implement}

We use Adam~\cite{DiederikPKingma2014Arxiv} for all optimization stages, with a learning rate of $0.001$ for pose optimization and $0.01$ for surface refinement. $\lambda_\text{skel} = 1.0$ and $\lambda_\text{kp} = 0.1$ balance the losses. We set $N_\text{samp}=5000$ for Chamfer distance computation. We initially render 18 views using three elevation levels with 4, 6, and 8 azimuths, respectively. The overlap ratio threshold $\tau_{\mathrm{IoU}}$ is set to $0.7$ to filter out views with poor initial alignment; this retains 17.75 views on average on FAUST and 15.92 views on average on TOPKIDS. For view selection, we set $N_\text{select} = 3$. The keypoints are predefined by the estimator--parametric-model interface. The human branch uses 70 MHR/SAM 3D Body joints. For the reported animal results, the frozen PRIMA estimator~\cite{yu2026prima} supplies SMAL parameters and projected SMAL joints.

For optional spectral-ICP refinement, we use the first $k=200$ LBO eigenvectors and run for $n_\text{iter} = 1$ iteration. This setting is not uniformly beneficial: the single step improves TOPKIDS, FAUST, and SCAPE, but worsens SMAL and SHREC19 in~\cref{tab:exp_non_isometric,tab:exp_isometric}. We therefore report it strictly as optional post-processing, separate from the primary pipeline; additional iterations tend to drift away from the original point-wise map and further degrade correspondence.

\section{Additional Robustness Results} \label{sec:suppl_robustness}

\begin{figure}[h]
    \centering
    \begin{subfigure}{0.47\linewidth}
        \centering
        \includegraphics[width=\linewidth]{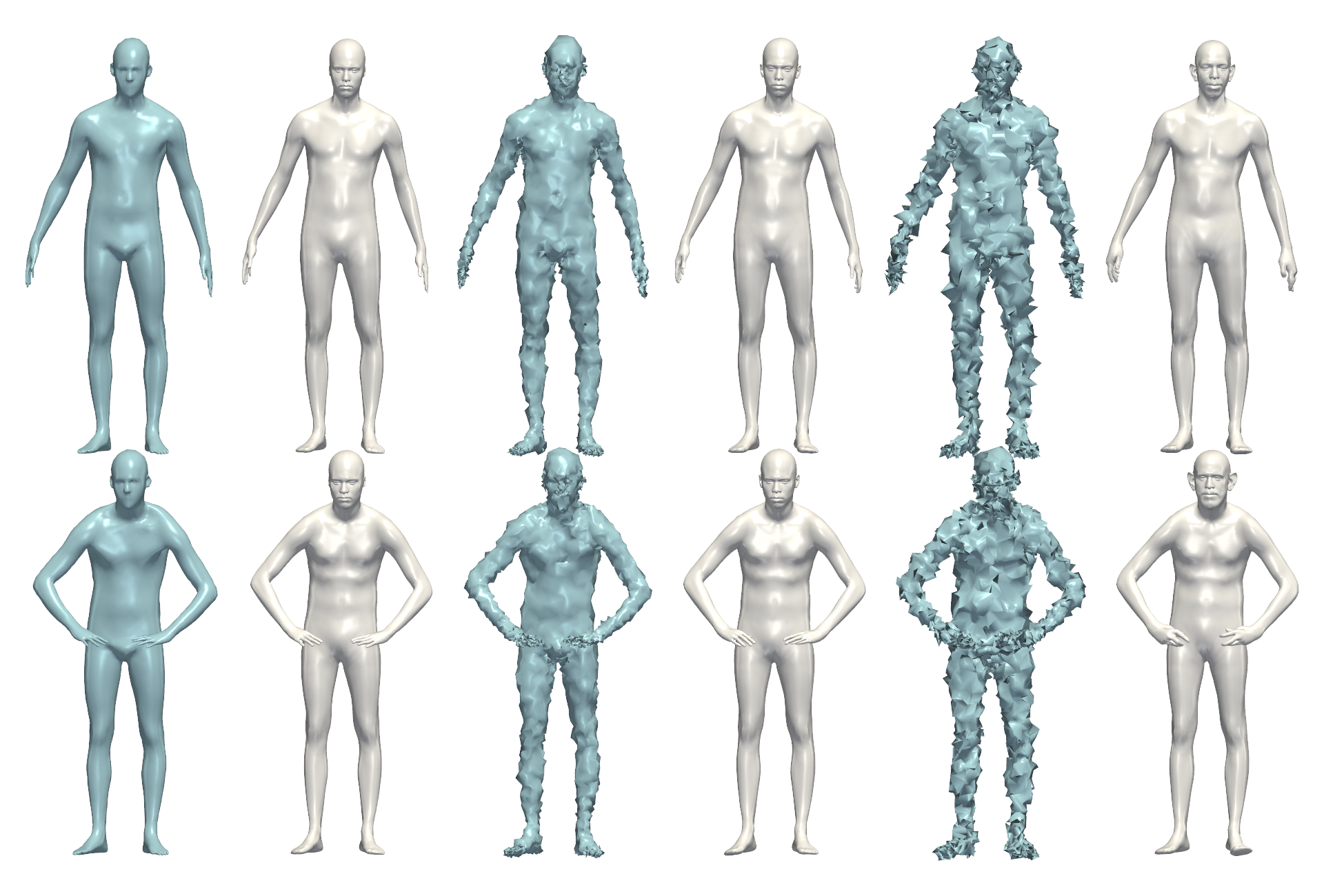}
        \caption{Noise robustness.}
    \end{subfigure}
    \hfill
    \begin{subfigure}{0.47\linewidth}
        \centering
        \includegraphics[width=\linewidth]{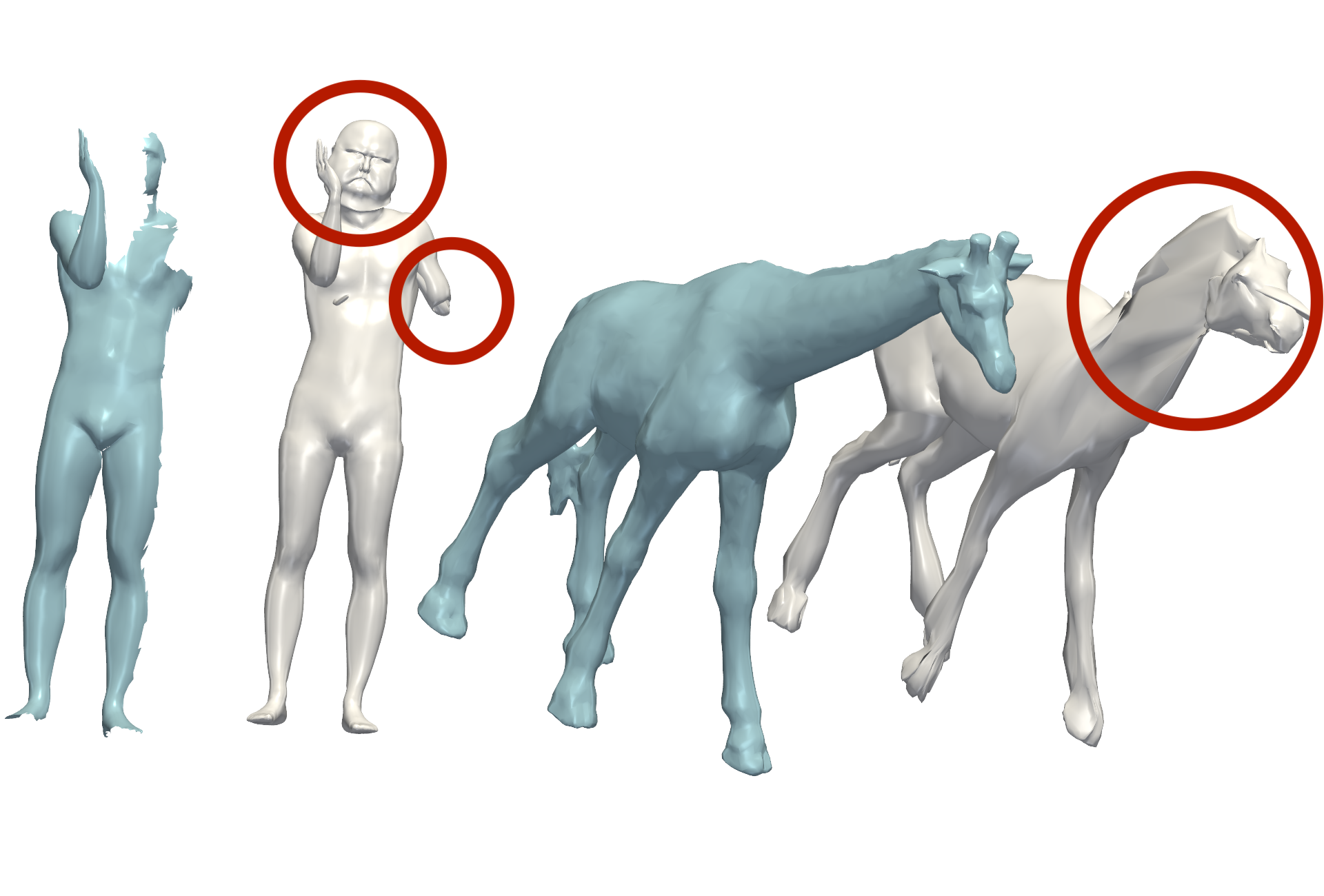}
        \caption{Failure cases.}
    \end{subfigure}
    \caption{Additional robustness analysis.}
    \label{fig:noise_failure}
\end{figure}

\noindent\textbf{Noise robustness.} We evaluate robustness to corrupted geometry by adding Gaussian noise to the input mesh vertices. As shown in~\cref{fig:noise_failure}, \ourmethod remains stable under substantial surface corruption because its core articulation is grounded through multi-view evidence rather than mesh connectivity.

\noindent\textbf{Failure cases.} \Cref{fig:noise_failure} also shows representative failure modes. The first is extreme truncation, where too many visible joints or surface cues are removed for reliable articulation. The second is applying the framework to a category (giraffe) poorly represented by the parametric prior and monocular estimator, which results in an appearance combining horse and camel features.

\begin{figure}[h]
    \centering
    \includegraphics[width=\linewidth]{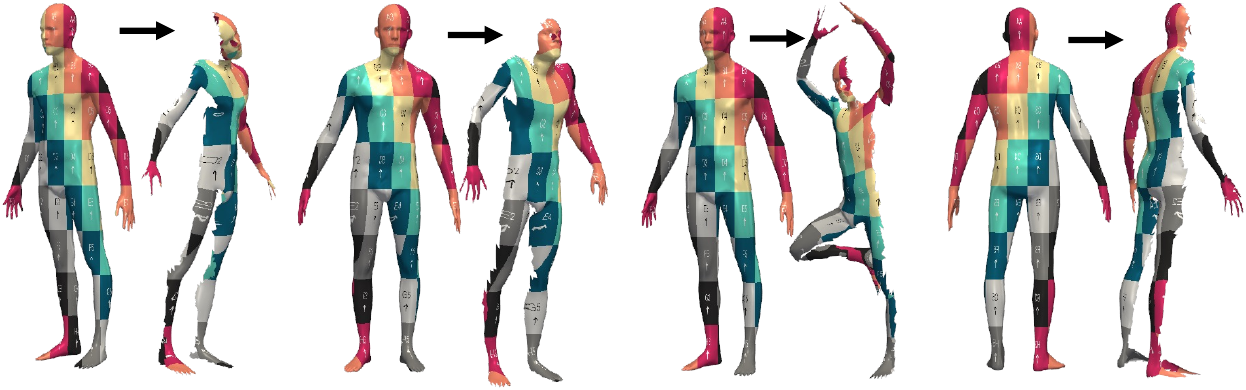}
    \caption{Selected texture-transfer examples on ray-cropped partial scans.}
    \label{fig:partial_scans}
\end{figure}

\section{Additional Animal Benchmarks and Prior Use} \label{sec:suppl_animals}

The name ``SMAL'' refers both to a parametric animal model and to a benchmark dataset. PRIMA~\cite{yu2026prima} uses the SMAL parametric model to annotate external animal data and synthesize animal shapes; the SMAL benchmark shapes used in the main paper are not directly used as PRIMA training instances. We use PRIMA's feed-forward prediction and do not invoke PRIMA's own parameter-updating test-time adaptation; only the per-view SMAL pose and shape variables are optimized.

Using the same PRIMA-based animal branch, we also evaluate on TOSCA and SHREC'20. As reported in~\cref{tab:suppl_animal}, \ourmethod has the lowest error among the evaluated methods on both datasets. The TOSCA difference from URSSM is small (6.89 versus 7.12), whereas the SHREC'20 difference is substantially larger (10.03 versus 34.18). URSSM is trained from scratch for 30 epochs.

\begin{table}[h]
    \centering
    \small
    \begin{tabular}{lcc}
        \toprule
        Method & TOSCA & SHREC'20 \\
        \midrule
        ZoomOut & 41.74 & 35.13 \\
        URSSM & 7.12 & 34.18 \\
        \ourmethod (ours) & \bfseries 6.89 & \bfseries 10.03 \\
        \bottomrule
    \end{tabular}
    \caption{Mean geodesic error $\times100$ ($\downarrow$) on additional animal-shape datasets. Values are aggregate point estimates.}
    \label{tab:suppl_animal}
\end{table}

\section{Runtime and High-Resolution Baseline Details} \label{sec:suppl_runtime}

\noindent\textbf{Runtime.} \ourmethod uses test-time optimization, so its uncached latency is higher than that of feed-forward or purely intrinsic methods. In the archived raw-FAUST sweep, articulating the 20 test shapes takes 21.63--23.05 minutes in total, or approximately 64.9--69.1 seconds per shape. Cached map composition ranges from 0.024 to 3.33 seconds per directed pair as resolution grows from 5k to raw. These quantities use different amortization units, as stated in~\cref{fig:scalability}. Possible accelerations include stopping optimization once its acceptance criteria are met, reusing parameters to initialize consecutive scans, and distilling \ourmethod's output into a faster 3D-native correspondence network.

\noindent\textbf{Spectral refinement and OOM.} The main matching signal of \ourmethod is the primary map obtained through shared parametric indices; spectral ICP is optional. The no-LBO primary pipeline reported at raw resolution is 2.11. A separate optional spectral-ICP run reached 1.93, but it is excluded from the scaling table because it introduces expensive spectral preprocessing without substantial accuracy gain.

The main high-resolution memory bottleneck in the evaluated spectral baselines is functional-to-pointwise-map conversion, which can materialize a dense $|V_x| \times |V_y|$ distance matrix. \ourmethod avoids this allocation through sparse assignment storage. For fair comparison, the main table reports a ZoomOut variant in which nearest search is evaluated in chunks to avoid OOM. This engineering change avoids the dense conversion allocation, but its error remains high and its pairwise runtime grows rapidly, as detailed in~\cref{tab:suppl_zoomout_chunked}.

\begin{table}[h]
    \centering
    \small
    \begin{tabular}{lcc}
        \toprule
        Resolution & Error & Time \\
        \midrule
        5k & 20.92 & 4m23s \\
        10k & 20.54 & 4m31s \\
        20k & 20.24 & 8m25s \\
        40k & 20.57 & 31m45s \\
        80k & 20.31 & 2h6m \\
        120k & 20.21 & 4h46m \\
        Raw & 19.89 & 11h6m \\
        \bottomrule
    \end{tabular}
    \caption{Chunked nearest-search ZoomOut on the 400 directed raw-FAUST pair queries. Error is the template-domain mean geodesic error $\times100$ ($\downarrow$); time is total wall-clock time over all queries.}
    \label{tab:suppl_zoomout_chunked}
\end{table}

At raw resolution, one uncached \ourmethod pair requires two articulations plus one composition, approximately $2(69.1)+3.33=141.5$ seconds, whereas chunked ZoomOut takes about $99.9$ seconds per pair. Caching reverses this comparison for repeated queries: processing the 20 test shapes and all 400 directed queries takes approximately 45.3 minutes for \ourmethod (23.05 minutes of articulation plus 22.2 minutes of composition), compared with 666 minutes for chunked ZoomOut. Thus, the measured advantage is collection-level amortization rather than lower cold-start latency for every single pair.

We cannot apply the same engineering tweak to URSSM because its loss computation relies on a full dense correspondence matrix. For the high-resolution raw-scan experiments in the main paper, URSSM is retrained from scratch for 30 epochs at each resolution, with Test-Time Adaptation disabled.

\section{Memory Efficient Geodesic Distance Evaluation} \label{sec:geodesic-eval}

The standard target-domain evaluation starts from template vertices and measures geodesic distances on the target shape, as shown in~\cref{fig:geodesic_benchmark_t2s}. For a directed pair $m\!\rightarrow\!n$, let $\hat\pi_{m\to n}:V_m\to V_n$ be the predicted map. For each shape $s\in\{m,n\}$, let $\tau_s:V_{\mathcal T}\to V_s$ denote its ground-truth template-to-shape map. Given an evaluated template query $t\in\Omega_{\mathcal T}\subseteq V_{\mathcal T}$, we first map it to the source vertex $\tau_m(t)$ and then apply the prediction. The predicted and ground-truth target vertices are
\begin{equation}
    n_{\mathrm{pred}}(t)
    =
    \hat\pi_{m\to n}\!\left(\tau_m(t)\right),
    \qquad
    n_{\mathrm{gt}}(t)
    =
    \tau_n(t).
    \label{eq:target-labels}
\end{equation}
The standard target-domain score used outside the high-resolution experiment is
\begin{equation}
    E_{\mathrm{target}}(m,n)
    =
    \frac{1}{|\Omega_{\mathcal T}|}
    \sum_{t\in\Omega_{\mathcal T}}
    \frac{D_n\!\left(n_{\mathrm{pred}}(t),n_{\mathrm{gt}}(t)\right)}
         {\sqrt{\operatorname{area}(n)}} ,
    \label{eq:eval-geodesic}
\end{equation}
where $D_n\in\mathbb R^{|V_n|\times|V_n|}$ is the target geodesic distance matrix and $\operatorname{area}(n)$ is the target surface area. Dataset scores average the directed test-pair scores uniformly.

Storing dense geodesic matrices for high-resolution target shapes is costly in both memory and disk space. A 200k $\times$ 200k matrix contains $4\times10^{10}$ entries, requiring 160~GB in single precision or 320~GB in double precision per target shape. This is hard to store to disk let alone load into memory.

\begin{figure}[h]
    \centering
    \begin{subfigure}{\linewidth}
        \centering\begin{tikzpicture}[
            align=center,
            node distance=1.5cm,
            box/.style={draw, rectangle, minimum width=1.8cm, minimum height=0.8cm, align=center},
            arrow/.style={->, >=Stealth, thick},
            scale=0.8, transform shape
        ]
            \node[box] (m1) {Shape $m$};
            \node[box, right=of m1] (n1) {Shape $n$};
            \node[box, above=of m1, xshift=1.5cm] (t1) {Template};
            \draw[arrow] (t1) -- node[left] {\small $\tau_m$} (m1);
            \draw[arrow] (m1) -- node[above] {\small $\hat\pi_{m\to n}$} (n1);
            \draw[arrow, red] (t1) -- node[right] {\small $\tau_n$} (n1);
            \node[draw=black, dashed, fill=gray!10, below=0.3cm of n1] {\small Query $D_n(n_{\mathrm{pred}}(t), n_{\mathrm{gt}}(t))$};
        \end{tikzpicture}
        \caption{}
        \label{fig:geodesic_benchmark_t2s}
    \end{subfigure}

    \begin{subfigure}{\linewidth}
        \centering\begin{tikzpicture}[
            align=center,
            node distance=1.5cm,
            box/.style={draw, rectangle, minimum width=1.8cm, minimum height=0.8cm, align=center},
            arrow/.style={->, >=Stealth, thick},
            scale=0.8, transform shape
        ]
            \node[box] (m2) {Shape $m$};
            \node[box, right=of m2] (n2) {Shape $n$};
            \node[box, above=of m2, xshift=1.5cm] (t2) {Template};
            \draw[arrow] (m2) -- node[above] {\small $\hat\pi_{m\to n}$} (n2);
            \draw[arrow] (n2) -- node[right] {\small $\gamma_n$} (t2);
            \draw[arrow, red] (m2) -- node[left] {\small $\gamma_m$} (t2);
            \node[draw=black, dashed, fill=gray!10, above=0.3cm of t2] {\small Query $D_t(t_{\mathrm{pred}}(x), t_{\mathrm{gt}}(x))$};
        \end{tikzpicture}
        \caption{}
        \label{fig:geodesic_benchmark_s2t}
    \end{subfigure}
    \caption{Comparison of geodesic distance evaluation approaches. (a) Template queries are mapped to the source and target via ground-truth maps \(\tau_m,\tau_n\), and errors are measured on the target. (b) Source-shape queries are evaluated in the shared template domain using the ground-truth shape-to-template maps \(\gamma_m,\gamma_n\).}
    \label{fig:geodesic_benchmark}
\end{figure}

To reduce both memory and disk usage in the high-resolution experiment, we instead evaluate in the template domain using ground-truth shape-to-template (s2t) correspondence, as shown in~\cref{fig:geodesic_benchmark_s2t}. This requires only one shared geodesic matrix $D_t\in\mathbb R^{|V_{\mathcal T}|\times|V_{\mathcal T}|}$ on an approximately 6k-vertex A-pose template. For 6,000 vertices, this matrix occupies only 288~MB in double precision and is reused for all shapes. For each shape $s\in\{m,n\}$, let $\gamma_s:V_s\rightarrow V_{\mathcal T}$ denote its archived ground-truth s2t map. Here, the evaluated query set $\Omega_m\subseteq V_m$ lies on the source shape. For each query $x\in\Omega_m$, the predicted and ground-truth template vertices are
\begin{equation}
    t_{\mathrm{pred}}(x)
    =
    \gamma_n\!\left(\hat\pi_{m\to n}(x)\right),
    \qquad
    t_{\mathrm{gt}}(x)
    =
    \gamma_m(x).
    \label{eq:template-labels}
\end{equation}
The directed-pair score is
\begin{equation}
    E_{\mathrm{temp}}(m,n)
    =
    \frac{1}{|\Omega_m|}
    \sum_{x\in\Omega_m}
    \frac{D_t\!\left(t_{\mathrm{pred}}(x),t_{\mathrm{gt}}(x)\right)}
         {\sqrt{\operatorname{area}(\mathcal T)}} .
    \label{eq:template-geodesic}
\end{equation}
We average~\cref{eq:template-geodesic} uniformly over the same 400 directed pair queries for every method. This metric measures error on the template domain and is therefore used only for within-table comparisons in~\cref{tab:exp_wild_scans}; we do not compare its absolute values with target-domain scores elsewhere.

\begin{algorithm}[h]
    \caption{Memory Efficient Geodesic Distance Evaluation}
    \label{alg:geodesic_eval}
    \begin{algorithmic}[1]
    \Require Prediction $\hat\pi_{m\to n}$ and query set $\Omega_m$
    \Require Ground-truth maps $\gamma_m:V_m\to V_{\mathcal T}$ and $\gamma_n:V_n\to V_{\mathcal T}$, template geodesic matrix $D_t$
    \ForAll{$x\in\Omega_m$}
        \State $t_{\mathrm{pred}}(x)\gets\gamma_n(\hat\pi_{m\to n}(x))$
        \State $t_{\mathrm{gt}}(x)\gets\gamma_m(x)$
    \EndFor
    \State Query $D_t(t_{\mathrm{pred}}(x), t_{\mathrm{gt}}(x))$ for each $x\in\Omega_m$
    \State Normalize by $\sqrt{\operatorname{area}(\mathcal T)}$ and average over $\Omega_m$
    \Ensure Template-domain error $E_{\mathrm{temp}}(m,n)$
    \end{algorithmic}
\end{algorithm}

This memory-efficient metric is applied only to the high-resolution raw-scan experiment. All other experiments use the target-domain score in~\cref{eq:eval-geodesic}. For both metrics, reported results are multiplied by $100$ after averaging over test pairs.

\section{Limitation}
\label{sec:limitations}

While \ourmethod demonstrates strong performance across varied triangle-mesh matching conditions, it has several limitations. First, the current formulation assumes input geometry is rendering compatible. Second, our method relies on a category-specific parametric model (\eg, MHR for humans or SMAL for animals) and a compatible single-view estimator, so extension to a new category requires both priors. Third, the proxy is bounded by the parametric model's expressiveness and may not preserve fine clothing, hair, or unusual proportions. Fourth, test-time optimization has higher cold-start latency than feed-forward inference; caching amortizes this cost only across repeated pair queries.

\end{document}